\documentclass[10pt,twocolumn,letterpaper]{article}
\newcommand{\ARXIV}[2]{#1} 

\newcommand{\code}[0]{\url{https://github.com/drprojects/DeepViewAgg}}
\newcommand{\wandb}[0]{\url{https://wandb.ai/damien_robert/DeepViewAgg-benchmark}}

\ARXIV{\usepackage[pagenumbers]{cvpr}}{\usepackage{cvpr}}

\makeatletter
\@namedef{ver@everyshi.sty}{}
\makeatother

\PassOptionsToPackage{hyphens}{url}
\usepackage[pagebackref=true,breaklinks=true,colorlinks,bookmarks=false]{hyperref}

\usepackage{times}
\usepackage{epsfig}
\usepackage{graphicx}
\usepackage{amsmath}
\usepackage{amssymb}
\usepackage{times}
\usepackage{graphicx}
\usepackage{amsmath}
\usepackage{amssymb}
\usepackage{booktabs}
\usepackage{multirow}
\usepackage{xcolor}
\usepackage{lipsum}
\usepackage{balance}
\usepackage{placeins}
\usepackage{subcaption}
\usepackage{algorithm}
\usepackage{algpseudocode}
\usepackage{tikz}
\usepackage{pgfplots}
\usepackage{styles}
\usepackage{efbox}
\usepackage{microtype}
\usetikzlibrary{shapes}
\usetikzlibrary{plotmarks}
\usepackage[accsupp]{axessibility}  
\efboxsetup{linecolor=white,linewidth=0.5pt,margin=-.25pt}

\def\cvprPaperID{2076} 
\def\confName{CVPR}
\def\confYear{2022}

\begin{document}

\title{Learning Multi-View Aggregation In the Wild\\ for Large-Scale 3D Semantic Segmentation}

\author{Damien Robert\textsuperscript{1, 2}\\
{\tt\small damien.robert@ign.fr}
\and
Bruno Vallet\textsuperscript{2}\\
{\tt\small bruno.vallet@ign.fr}

\and
Loic Landrieu\textsuperscript{2}\\
{\tt\small loic.landrieu@ign.fr}

\and
{\textsuperscript{1}CSAI, ENGIE Lab CRIGEN, Stains, France}\\
{\textsuperscript{2}Univ Gustave Eiffel, ENSG, IGN, LASTIG, F-77454 Marne-la-Vallee, France}\\
}

\maketitle

\begin{abstract}
Recent works on 3D semantic segmentation propose to exploit the synergy between images and point clouds by processing each modality with a dedicated network and projecting learned 2D features onto 3D points. Merging large-scale point clouds and images raises several challenges, such as constructing a mapping between points and pixels, and aggregating features between multiple views. Current methods require mesh reconstruction or specialized sensors to recover occlusions, and use heuristics to select and aggregate available images. In contrast, we propose an end-to-end trainable multi-view aggregation model leveraging the viewing conditions of 3D points to merge features from images taken at arbitrary positions. Our method can combine standard 2D and 3D networks and outperforms both 3D models operating on colorized point clouds and hybrid 2D/3D networks without requiring colorization, meshing, or true depth maps. We set a new state-of-the-art for large-scale indoor/outdoor semantic segmentation on S3DIS ($74.7$ mIoU $6$-Fold) and {on KITTI-360 ($58.3$ mIoU)}. Our full pipeline is accessible at \code, and only requires raw 3D scans and a set of images and poses.
\end{abstract}
\setlength{\parskip}{-0.13em}
\section{Introduction}
The fast-paced development of dedicated neural architectures for 3D data has led to significant improvements in the automated analysis of large 3D scenes \cite{guo2020deep}.
All top-performing methods operate on colorized point clouds, which requires either specialized sensors \cite{woodhouse2011multispectral}, or running a colorization step which is often closed-source \cite{trimble,faro,leica} and sensor-dependent \cite{julin2020evaluating}.
However, while colorized point clouds carry some radiometric information, images combined with dedicated 2D architectures are better suited for learning textural and contextual cues.
A promising line of work sets out to leverage the complementarity between 3D point clouds and images by projecting onto 3D points the 2D features learned from real {\cite{dai20183dmv,hu2021bidirectional, jaritz2019multi}} or virtual images \cite{kundu2020virtual, chiang2019unified}

\begin{figure}[t]
    \centering
    \includegraphics[width=\columnwidth]{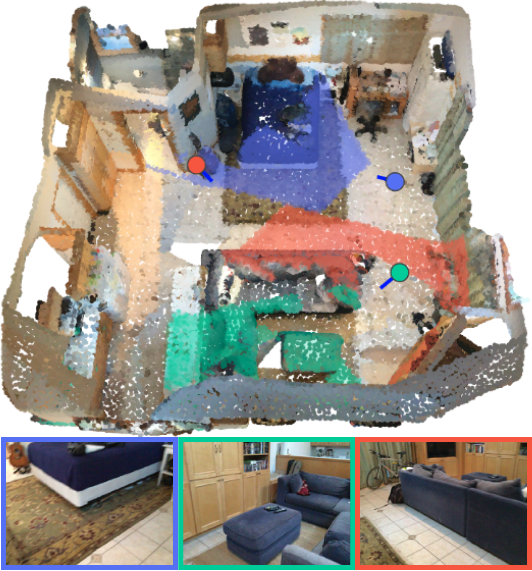}
    \caption{
    {\bf Combining 2D and 3D Information.} 
    We propose to merge the complementary information between point clouds and a set of co-registered images. Using a simple visibility model, we can project 2D features onto the 3D points and use viewing conditions to select features from the most relevant images.
    We represent images at their position with the symbol~ 
    \protect\begin{tikzpicture}
    \protect\draw[draw=blue, very thick] (0.0,0.0) -- (0.3,0.0);
    \protect\fill[draw=black!70!white, thick, fill=cyan, scale=1] (0.0,0.0) circle (0.1);
    \protect\end{tikzpicture}
    and color the 3D points according to the image they are seen in.
    }
    \label{fig:teaser}
    \vspace{-5mm}
\end{figure}
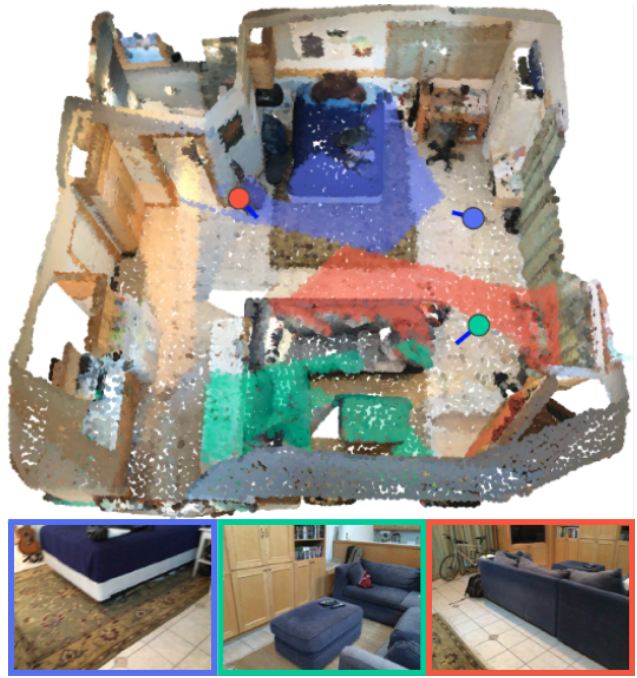

Combining point clouds and images with arbitrary poses (\ie \emph{in the wild}) as represented in \figref{fig:teaser}, involves recovering occlusions and computing a point-pixel mapping, which is typically done using accurate depth maps from specialized sensors \cite{valkenburg1998accurate,cai2017rgb} or a potentially costly meshing step \cite{boulch2018snapnet}.
Furthermore, when a point is seen in different images simultaneously, the 2D features must be merged in a meaningful way.
In the mesh texturation literature, multi-view aggregation is typically addressed by selecting images for each triangle based on their viewing conditions, \eg distance, viewing angle, or occlusion \cite{ allene2008seamless, lempitsky2007seamless, waechter2014let}.
Hybrid 2D/3D methods for large-scale point cloud analysis {usually} rely on heuristics to select a fixed number of images per point and pool their features uniformly without considering viewing conditions.
Multi-view aggregation has also been extensively studied for shape recognition \cite{feng2018gvcnn,su2015multi,wei2020view}, albeit in a controlled and synthetic setting not entirely applicable to the analysis of large scenes.

In this paper, we propose to learn to merge features from multiple images with a dedicated attention-based scheme.
For each 3D point, the information from relevant images is aggregated based on the point's viewing condition.
Thanks to our GPU-based implementation, we can efficiently compute a point-pixel mapping without mesh or true depth maps, and without sacrificing precision.
Our model can handle large-scale scenes with an arbitrary number of images per point taken at any position (with camera pose information), which corresponds to a standard industrial operational setting \cite{hodgetts2013laser, virtanen2020interactive, pepe20163d}.
Using only standard 2D and 3D backbone networks, we set a new state-of-the-art for the S3DIS and KITTI-360 datasets.
Our method improves on both standard and hybrid 2D/3D approaches without requiring point cloud colorization, mesh reconstruction, or depth sensors. 
In this paper, we present a novel and modular multi-view aggregation method for semantizing hybrid 2D/3D data based on the viewing conditions of 3D points in images. Our approach combines the following advantages:
\begin{itemize}
    \item  {We set a new state-of-the-art for S3DIS 6-fold ($74.7$ mIoU), and KITTI-360 Test ($58.3$ mIoU) without using points' colorization.}
    \item {Our point-pixel mapping operates directly on 3D point clouds and images without requiring depth maps, meshing, colorization, or virtual view generation.}
    \item {Our efficient GPU-based implementation handles arbitrary numbers of 2D views and large 3D point clouds.}
\end{itemize}
\section{Related Work}
\paragraph{Attention-Based Modality Fusion.}
Methods using attention mechanisms to learn 
multi-modal representation have attracted a lot of attention, in particular for combining textual and visual information \cite{caglayan2016multimodal, huang2016attention, gu2018multimodal} as well as videos \cite{hori2017attention, long2018multimodal}.
Closer to our setting, Lu \etal  \cite{lu2016hierarchical} use an attention scheme to select the most relevant parts of an image for visual question answering.
Li \etal \cite{li2020attention} define a two-branch attention-based modality fusion network merging 2D semantic and 3D occupancy for scene completion.
Such work confirms the relevance of using attention for learning multi-modal representations.

\paragraph{2D/3D Scene Analysis with Deep Learning.}
Over the last few years, deep networks specifically designed to handle the 3D modality have reached impressive degrees of performance and maturity, see the review of Guo \etal \cite{guo2020deep}. 
Recent work \cite{dai20183dmv,jaritz2019multi,hu2021bidirectional} propose to use a dedicated 3D network for processing point clouds, while a 2D convolutional network extracts radiometric features which are projected to the point cloud. These methods require the true depth of each pixel to compute the point-pixel mapping, which makes them less applicable in a real-world setting. SnapNet \cite{boulch2018snapnet}, as well as more recent work
\cite{kundu2020virtual,chiang2019unified} generate virtual views processed by a 2D network and whose predictions are then projected back to the point cloud. These approaches, while performing well, require a costly mesh reconstruction preprocessing step to generate meaningful images.
Some approaches \cite{hazirbas2016fusenet, krispel2020fuseseg} fuse RGB and range images, which requires dedicated sensors and can not handle multiple views with occlusions.
Existing hybrid 2D/3D methods rely on a fixed number of images per point chosen with heuristics such as the maximization of unseen points~\cite{dai20183dmv, jaritz2019multi, kundu2020virtual}. Then, the different views are merged using pooling operations (max~\cite{su2015multi, dai20183dmv} or sum-pool~\cite{jaritz2019multi}) or based on the 2D features' content~\cite{hu2021bidirectional}. To the best of our knowledge, no method has yet been proposed to leverage the viewing conditions for multi-view aggregation for the semantic segmentation of large scenes.

\ARXIV{}{\vspace{-5mm}}
\paragraph{View Selection.} 
The problem of selecting and merging the best images for a 3D scene has been extensively studied for surface reconstruction and texturing. Images are typically chosen according to the viewing angle with the surface normal \cite{lempitsky2007seamless,birchfield1999multiway}, proximity and resolution \cite{allene2008seamless, buehler2001unstructured}, geometric and visibility priors \cite{schonberger2016pixelwise}, as well as \emph{crispness} \cite{gal2010seamless}, and consistency with respect to occlusions \cite{waechter2014let}. While most of these criteria  do not directly apply to point clouds, they illustrate the importance of camera pose information for selecting relevant images.

Related to our setting is the \emph{Next Best View} selection problem \cite{scott2003view}, which consists in planning the camera position giving the \emph{most information} about an object of interest \cite{connolly1985determination}. This criterion takes different meanings according to the setting, such as the number of unseen voxels \cite{vasquez2017view}, diversity \cite{mokhtarian2000automatic}, information-theoretic measures of uncertainty \cite{isler2016information}, or can be directly learned end-to-end \cite{mendoza2020supervised, wu20153d}. Our setting differs in that the images have already been acquired, and the task is to choose which one contains the most relevant information for each point.
We draw inspiration from the end-to-end approaches demonstrating that a neural network can assess the quality of information contained in an image from pose information.

The problem of view selection is also addressed in the literature on shape recognition \cite{su2015multi}. Features from different images can be merged based on their similarity \cite{wang2019dominant}, \emph{discriminativity} \cite{feng2018gvcnn}, or using patch matching schemes \cite{yang2019learning, yu2018multi} or graph-neural networks \cite{wei2020view}. Some methods use 3D features \cite{you2019pvrnet} or camera position \cite{kanezaki2018rotationnet, hamdi2021mvtn} to select the best views, but no technique yet makes explicit use of the viewing configuration. Furthermore, these methods operate on synthetic views of artificial shapes, which differs from our goal of analyzing large scenes with images in arbitrary poses.

{Closer to our problem, Armeni \etal \cite{armeni20193d} aggregate views using handcrafted heuristics. Bozic \etal \cite{bozic2021transformerfusion} use a distance-aware attentive view aggregation for 3D reconstruction, but disregard other viewing conditions. 
}

\section{Method}


Let $\PP$ be a set of 3D points and $\II$ a collection of co-registered images, all acquired from the same scene. We characterize points by their position in space, and images by their pixels' RGB values along with intrinsic and extrinsic camera parameters.
Our goal is to exploit the correspondence between points and image pixels to perform 3D point cloud semantic segmentation with features learned from both modalities.
Our method starts by computing an {occlusion-aware} mapping between 3D points and pixels, then uses viewing conditions through an attention scheme to aggregate relevant image features for each 3D point. This approach can be easily integrated into a standard 3D network architecture, allowing us to learn from both point clouds and images simultaneously in an end-to-end fashion.
%
\subsection{Point-Image Mapping}
\label{sec:mapping}
We start by efficiently computing a mapping between the images of $\II$ and the points of $\PP$.
We say that a point-image pair $(p,i) \in \PP \times \II$ is \emph{compatible} if $p$ is visible in $i$, \ie $p$ is in the frustum of $i$ and not occluded. For such a pair, we define the re-projection $\cM(p,i)$ as the pixel of $i$ in which $p$ is \emph{visible}. Note that as points are zero-dimensional objects (zero-volume), $\cM(p,i)$ is a single pixel. We denote by $v(p)$ the \emph{views} of $p$, \ie the set of images in which $p$ is visible.

\begin{figure}[t]
    \centering
    \begin{tabular}{cc}
        \begin{subfigure}{.45\columnwidth}
        \includegraphics[width=\columnwidth]{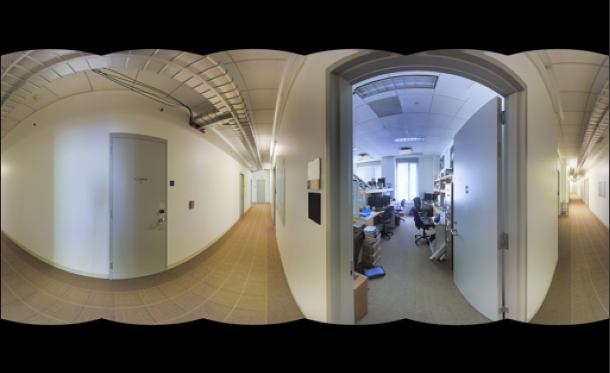}
        \caption{Image.}
        \label{fig:splat:image}
        \end{subfigure}
        &
        \begin{subfigure}{.45\columnwidth}
        \includegraphics[width=\columnwidth]{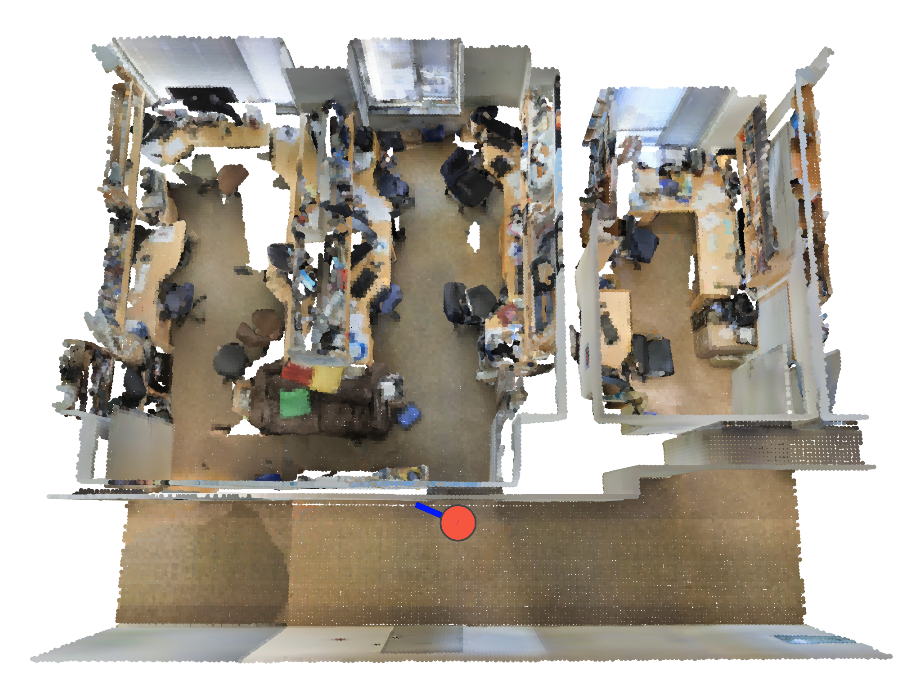}
        \caption{Point Cloud.}
        \label{fig:splat:pc}
        \end{subfigure}
        \\
        \begin{subfigure}{.45\columnwidth}
        \includegraphics[width=\columnwidth]{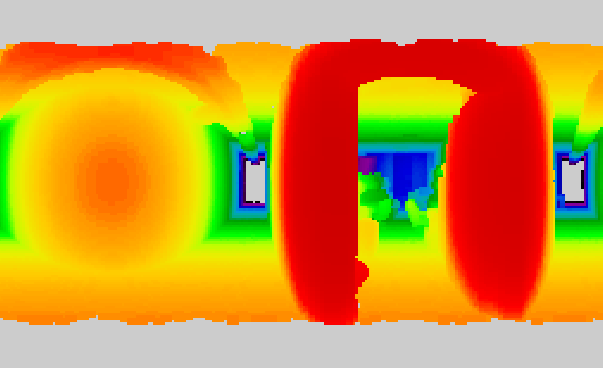}
        \caption{Z-buffer.}
        \label{fig:splat:splat}
        \end{subfigure}
        &
        \begin{subfigure}{.45\columnwidth}
        \includegraphics[width=\columnwidth]{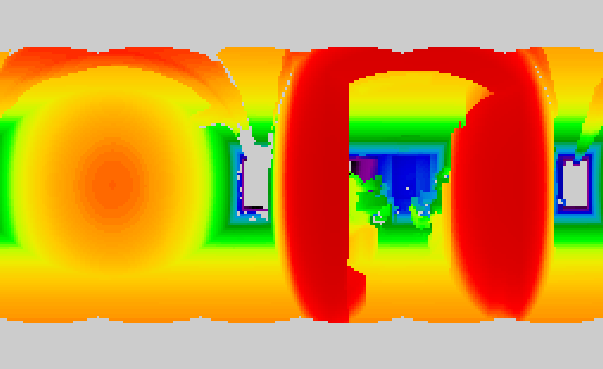}
        \caption{True depth.}
        \label{fig:splat:surface}
        \end{subfigure}
    \end{tabular}
    \caption{{\bf Mapping Computation.} We estimate pixel depth for all \Subref{fig:splat:image} images using the co-registered \Subref{fig:splat:pc} point cloud. We compute \Subref{fig:splat:splat} Z-buffers
     with an efficient GPU-accelerated implementation, resulting in depth maps comparable to the \Subref{fig:splat:surface} true distance given by camera-aligned depth sensors. We use our estimated depth maps to compute point-image mappings. Better seen on a monitor.}
    \label{fig:splat}
    \vspace{-5mm}
\end{figure}

\paragraph{Point-Pixel Mapping Construction.}
We operate in a general \emph{in the wild} multi-view setting in which the optical axes of the cameras and the 3D sensor are not necessarily aligned. Consequently, computing the point-image mapping requires a \emph{visibility model} to detect occlusions.
This can be done by computing a full mesh reconstruction from the point clouds or by using a depth map obtained by a camera-aligned depth sensor or other means. 
In contrast, we propose an efficient implementation of the straightforward Z-buffering method \cite{STRASSER1982105} to compute the mapping directly from images and point clouds.
For each image  $i \in \II$, we replace all 3D points in the frustum of $i$ under a pre-determined distance by a {square plane section} facing towards $i$ and whose size depends on their distance to the sensor and the resolution of the point cloud.
We can compute the projection mask---or \emph{splat}---of each square onto $i$ using the camera parameters of $i$.
We iteratively accumulate all splats in a depth map called {Z-buffer} by keeping track of the closest point-camera distance for each pixel. Simultaneously, we store corresponding point indices in an {index map}, along with other relevant point attributes. Once all splats have been accumulated, \emph{visible} points are the ones whose indices appear in the index map.
For each visible point $p$, we set $\cM(p,i)$ as the pixel of $i$ in which $p$ itself is projected.
{Our GPU-accelerated implementation can process the entire S3DIS dataset \cite{armeni20163d} subsampled at $5$cm ($12$ million points and $1413$ high-resolution equirectangular images) within $65$ seconds.}
See \figref{fig:splat} for an illustration, and the Appendix for more details on the mapping computation, alternative visibility models, and our memory-efficient implementation for large-scale mappings.

\paragraph{Viewing Conditions.} To each compatible point-image pair $(p,i)$, we associate a $D$-dimensional vector $o_{(p,i)}$ describing the conditions under which the point $p$ is seen in $i$.
In practice, we define this vector as a set of $D=8$ handcrafted features qualifying the observation conditions of $(p,i)$:
(i) normalized depth, 
(ii-iv) local geometric descriptors (linearity, planarity, scattering)
(v) viewing angle \wrt the estimated normal,
(vi) row of the pixel,
(vii) local density,
(viii) occlusion rate.
See the Appendix for more details on the computation and impact of these values.
\subsection{Learning Multi-View Aggregation}
\label{sec:view}
We denote by $\{\f{2}_i\}_{i \in \II}$ a set of 2D feature maps of width $C$ associated to the images $\II$, typically obtained with a convolutional neural network (CNN). Our goal is to transfer these features to the 3D points by exploiting the correspondence between points and images. However, not all viewing images contain equally relevant information for a given 3D point. We propose an attention-based approach to weigh and aggregate features from the viewing images for each point $p$.

\begin{figure}[ht]
    \centering
    \includegraphics[width=\columnwidth]{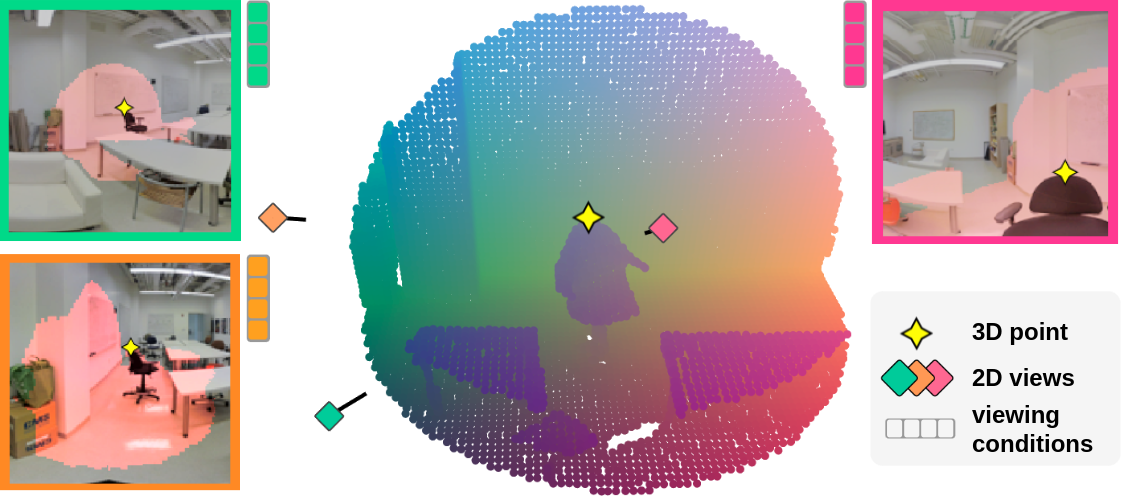}
    \definecolor{greentemp}{rgb}{0,0.85,0.53}
    \definecolor{orangetemp}{rgb}{1,0.53,0.14}
    \definecolor{pinktemp}{rgb}{1,0.22,0.57}
    \caption{
        {\bf Multi-View Information.}
        A 3D point \includegraphics[width=0.25cm]{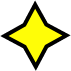} is seen in several images with different insights. Here, the \textcolor{greentemp}{\bf green} image contains contextual information, while the \textcolor{pinktemp}{\bf pink} image captures the local texture. The \textcolor{orangetemp}{\bf orange} image sees the point at a slanted angle and may {contain no additional relevant} information.
    }
    \label{fig:deepview}
    \ARXIV{}{\vspace{-3mm}}
\end{figure}

\paragraph{View Features.} The mapping $\cM(p,i)$ described in \secref{sec:mapping} allows us to associate image features to each compatible point-image pair $(p,i)$:
\begin{align}\label{eq:viewencoder}
  \ft{2}_{(p,i)} &= 
  \MLP
  \left( 
  \f{2}_i\left[\cM(p,i)\right]
  \right)~,
\end{align}
with $\MLP:\bR^C \mapsto \bR^C$ a Multi-Layer Perceptron (MLP).
Learned image features can contain information of different natures: contextual, textural, class-specific, and so on. To reflect this consideration, we split the channels of $\ft{2}_{(p,i)}$ into $K$ contiguous blocks of $\lfloor C/K\rfloor$ channels:
\begin{align}\label{eq:groups}
\ft{2}_{(p,i)}=\left[\ft{2}_{(p,i),1},\cdots,\ft{2}_{(p,i),K}\right]~.
\end{align}
with $[\,\cdot\,]$ the channel-wise concatenation operator. Each block of channels represents a subset of the image information contained in $\ft{2}$.
\paragraph{View Quality.}
The conditions under which a point is seen in an image can be more or less conducive to certain types of information, see \figref{fig:deepview}.
For example, an image viewing a point from a distance may give important contextual cues, while an image taken close and at a straight angle may give detailed textural information. In contrast, an image in which a point is seen from a slanted angle or under high distortion may not contain relevant information and may need to be discarded.
To model these complex dependencies, we propose to predict for each compatible point-image pair $(p,i)$ a set of $K$ \emph{quality scores} $x_{(p,i)}^k \in \bR$ from its viewing conditions $o_{(p,i)}$ defined in \secref{sec:mapping}. The quality $x_{(p,i)}^k$ represents the relevance for point $p$ of the information contained in the feature block $k$ of image $i$ .

For each point $p$, we consider the set $v(p)$ of images in which it is visible. We propose to learn to predict the view quality $x_{(p,i)}^k$ for each feature block $k$ by considering all images $i\in v(p)$ \emph{simultaneously}. Indeed, the relevance of an image can depend on the context of the other views. For example, while a given image may provide less-than-perfect viewing conditions of a given 3D point, it may be the only available image with global information of the point's context.
We use a deep set architecture \cite{zaheer2017deep} to map the set of viewing conditions $\{{o}_{(p,i)} \}_{i \in v(p)}$ to a vector of size $K$:
\begin{align}
    z_{(p,i)} &= \phi_1(o_{(p,i)})\\\label{eq:deepset}
    x_{(p,i)} &= \phi_3
    \left(
        \left[
            {z}_{(p,i)},
            \phi_2
            \left(
                \max
                    \{{z}_{(p,i)} \}_{i \in v(p)}
            \right)
        \right]
    \right)~,
\end{align}
with $\phi_1:\bR^D\mapsto \bR^M$, $\phi_2:\bR^M\mapsto \bR^M$, and $\phi_3:\bR^{2M} \mapsto \bR^K$ three MLPs, $M$ the size of the set embedding, and $\max$ the channelwise maximum operator for a set of vectors.

\paragraph{View Attention Scores.} We can now compute $K$ attention scores ${a}_{(p,i)}^k$ in $[0,1]$ corresponding to the relative relevance for point $p$ of the $k$th feature block of image $i$.
The attentions are obtained by applying a softmax function to the quality scores $x_{(p,i)}^k$ across the images in $v(p)$. To account for the possibly varying number of views per point, we scale the softmax according to the number of images seeing the point $p$ : 
\begin{align}
{a}_{(p,i)}^k = \softmax 
\left(
    \frac1{\sqrt{\vert v(p) \vert}} 
    \left\{
        x_{(p,i)}^k
    \right\}_{i \in v(p)}
\right)~.
\end{align}
\ARXIV{}{\vspace{-0.5cm}}
\paragraph{View Gating.} A limitation of using a softmax in this context is that the attention scores $\tilde{a}_{(p,i)}^k$ always sum to $1$ over $v(p)$ regardless of the overall quality of the image set. Because of occlusion or limited viewpoints, some 3D points may not be seen by any relevant image for a given feature block $k$ (\eg no close or far images). In this case, it may be beneficial to discard an information block from all images altogether and purely rely on geometry. This allows the 2D network to learn image features without accounting for potentially spreading corrupted information to points with dubious viewing conditions.
To this end, we introduce a gating parameter $g_p^k$ whose role is to block the transfer of the features block $k$ if the overall quality of the image set $v(p)$ is too low:
\begin{align}\label{eq:gating}
&g_{p}^k = 
\relu \left(
\tanh
    \left( 
        \alpha_k
        \max_{i \in v(p)}
        \left(
            x_{(p,i)}^k
        \right)
        +\beta_k
    \right)
\right)
~,
\end{align}
with $\alpha, \beta\in \bR^K$ trainable parameters and $\relu$ the rectified linear activation \cite{nair2010rectified}. If all quality scores $x_{(p,i)}^k$ are negative for a given point $p$ and block $k$, the gating parameter $g_{p}^k$ will be exactly zero and block possibly detrimental information due to sub-par viewing conditions.

\paragraph{ Attentive Image Feature Pooling.}
For each point $p$ seen in one or more images, we merge the feature maps $\ft{2}_{(p,i)}$ from each view $(p,i)$.
For each block $k$, we compute the sum of the view features $\ft{2}_{(p,i),k}$ weighted by their respective attention scores $a^k_{(p,i)}$ and multiplied by the gating parameter $g_p^k$. The combined image feature $\cP(\f{2},p)$ associated to point $p$ is then defined as the channelwise concatenation of the resulting tensors for all blocks:
\begin{align}
  \cP(\f{2},p)  = 
  \left[
  g_p^k
      \sum_{i \in v(p)}
      a^k_{(p,i)}
      \ft{2}_{(p,i),k}
  \right]_{k=1}^K~.
\end{align}

\begin{figure}[t]
    \centering
    \includegraphics[width=0.98\columnwidth]{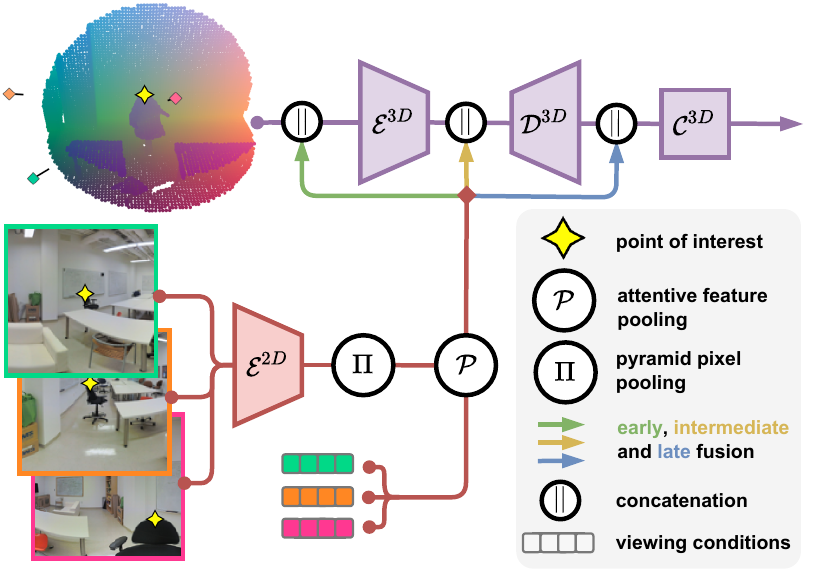}
    \caption{{\bf Bimodal 2D/3D Architecture.} Using our multi-view aggregation module, we combine a 2D convolutional encoder $\EncTWO$ and a 3D network composed of an encoder $\EncTHREE$, a decoder $\DecTHREE$, and a classifier $\ClassTHREE$. We associate relevant 2D features to each 3D point according to their viewing conditions in each compatible image. We propose three different 2D/3D fusion strategies: early (our choice in the experiments), intermediate, and late fusion.}
    \label{fig:bimodal}
    \ARXIV{}{\vspace{-5mm}}
\end{figure}

\subsection{Bimodal Point-Image Network.}
We can use the multi-view feature aggregation method described above to perform semantic segmentation of a point cloud and co-registered images by combining a network operating on 3D point clouds and 2D CNN.

\paragraph{Fusion Strategies.}
We use a 2D fully convolutional network to compute pixel-wise image feature maps $\f{2}$.
We also consider a 3D deep network following the classic $U$-Net architecture \cite{ronneberger2015u} and composed of three parts: (i) an encoder  $\EncTHREE$ mapping the point cloud into a set of 3D feature maps at different resolution (\emph{innermost} map and skip connections); (ii) a decoder $\DecTHREE$ converting these maps into a 3D feature map at the highest resolution (iii) a classifier $\ClassTHREE$ associating to each point a vector for class scores of size $N$, the number of target classes.

\begin{figure}[t]
    \centering

    \begingroup
    \setlength{\tabcolsep}{0pt}
    \def\arraystretch{0.0}
    \begin{tabular}{ccccc}
    \multicolumn{2}{c}{
     \efbox{\includegraphics[width=.48\linewidth]{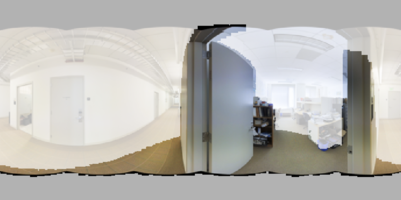}}
     }
     &
     \multicolumn{3}{c}{
     \efbox{\includegraphics[width=.48\linewidth]{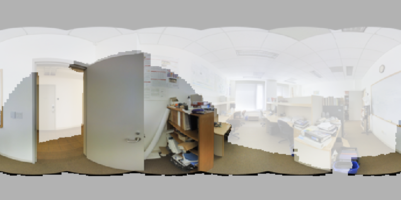}}
     }
     \\
    \begin{minipage}[t][.24\linewidth][t]{.24\linewidth}
    \efbox{\includegraphics[width=1\linewidth]{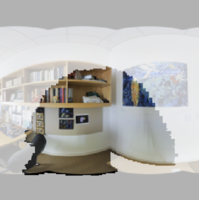}}
    \end{minipage}
    &
    \begin{minipage}[t][.24\linewidth][t]{.24\linewidth}
    \efbox{\includegraphics[width=1\linewidth]{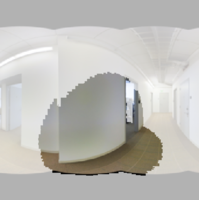}}
    \end{minipage}
    &
    \begin{minipage}[t][.24\linewidth][t]{.24\linewidth}
    \efbox{\includegraphics[width=1\linewidth]{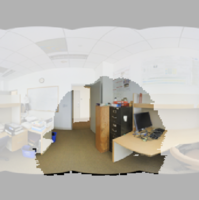}}
    \end{minipage}
    &
    \begin{minipage}[t][.24\linewidth][t]{.24\linewidth}\vspace{-2.02cm}
    \begin{tabular}{ll}
    \efbox{\includegraphics[width=.49\linewidth]{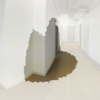}}
    &
    \efbox{\includegraphics[width=.49\linewidth]{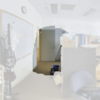}}
    \\
    \efbox{\includegraphics[width=.49\linewidth]{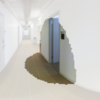}}
    &
    \begin{minipage}[t][.5\linewidth][t]{.5\linewidth}\vspace{-1.01cm}
    \begin{tabular}{l}
    \efbox{\includegraphics[width=.98\linewidth]{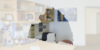}}\\
    \efbox{\includegraphics[width=.98\linewidth]{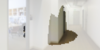}}
    \end{tabular}
    \end{minipage}
    \end{tabular}
    \end{minipage}
    \end{tabular}
    \endgroup
    \vspace{-2cm}
    \caption{{\bf Dynamic Batching.} We can improve the quantity of information contained in each training batch by cropping images around the sampled point clouds. We represent a set of $10$ images with different crop size fitting in a budget of pixels corresponding to $4$ full-size images.}
    \label{fig:dynamicbatching}
\end{figure}

As shown in \figref{fig:bimodal}, we investigate three classic fusion schemes \cite{hazirbas2016fusenet, jaritz2019multi, krispel2020fuseseg}, connecting the image features at different points of the 3D network:
(i) directly with the raw 3D features before $\EncTHREE$ (\emph{early fusion}), 
(ii) in the skip connections (\emph{intermediate fusion})
(iii) between the decoder $\DecTHREE$ and the classifier $\ClassTHREE$ (\emph{late fusion:}). See the Appendix for the details and equations for these fusion schemes.
%
%
%
\paragraph{Dynamic-Size Image-Batching}
The number of images $v(p)$ in which a point $p$ is visible can vary significantly. Furthermore, when dealing with large-scale scenes, only a subset $\PPsample$ of the 3D scene is typically processed at once (\eg spherical sampling). For this reason, the part of an image $i$ for which points of $\PPsample$ are visible can sometimes be only a small fraction of the entire image. This will typically occur with equirectangular images or when $\PPsample$ is far away from $i$. We use the adaptive batching scheme depicted in \figref{fig:dynamicbatching} to stabilize memory usage across batches and avoid needless computations on excessively large images. The first step is to crop each image using the smallest window across a fixed set of sizes (\eg $64\times 64$, $128\times 64$, etc.) such that the crop contains the bounding box of all seen points of $\PP$ with a given margin. Observing that the memory consumption of a fully convolutional encoder is linear \wrt the number of input pixels, we allocate to each point cloud in the batch a \emph{budget} of pixels.
Images are then chosen randomly by iteratively selecting images with a probability proportional to their number of pixels and to the number of newly seen points in the cloud, until the pixel budget is spent. Finally, the images are organized into different batches according to their sizes, allowing for their simultaneous processing.
Note that at inference time, we can take batches as large as the GPU memory allows.
\addtolength{\tabcolsep}{-3pt}
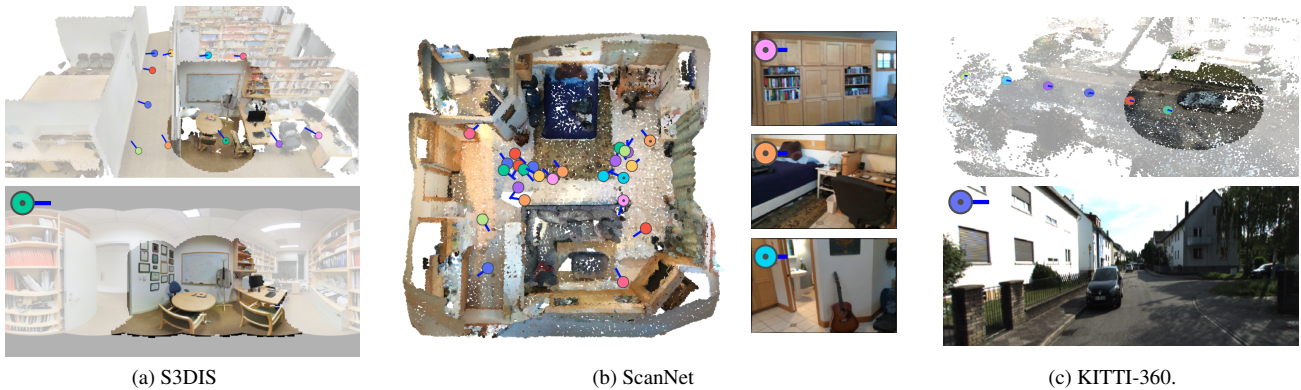
\begin{figure*}[ht!]
\centering
   \begin{tabular}{cccc}
   \begin{subfigure}{0.27\linewidth}
   \begin{tabular}{c}
    \begin{tikzpicture}
        \node[anchor=south west,inner sep=0] (image) at (0,0) {\includegraphics[width=1\linewidth, height=0.48\linewidth]{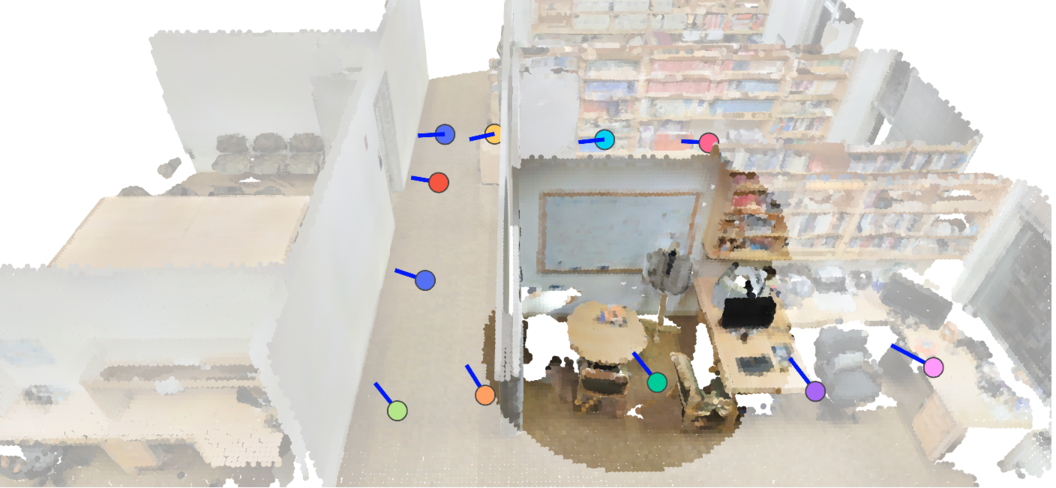}};
        \begin{scope}[x={(image.south east)},y={(image.north west)}]
        \fill[draw=none,fill=black!70!white, scale=1] (0.622,0.2165) circle (0.15mm);
        \end{scope}
 \end{tikzpicture}
   \\
   \begin{tikzpicture}
        \node[anchor=south west,inner sep=0] (image) at (0,0) {\includegraphics[width=1\linewidth, height=0.48\linewidth]{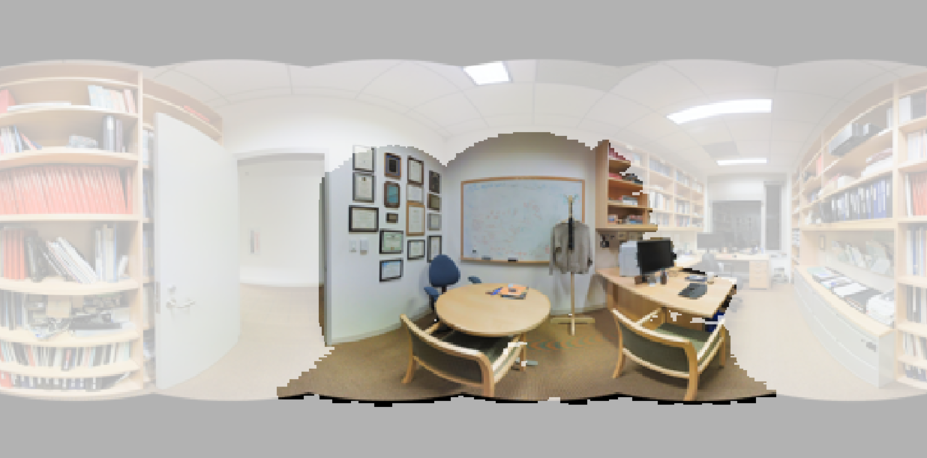}};
        \begin{scope}[x={(image.south east)},y={(image.north west)}]
        \draw[draw=blue, ultra thick] (0.05,0.9) -- (0.13,0.9);
        \definecolor{tempcolor}{rgb}{0,0.8,0.6}
        \fill[draw=black!70!white, thick, fill=tempcolor, scale=1] (0.05,0.9) circle (1.5mm);
        \fill[draw=none,fill=black!70!white, scale=1] (0.05,0.9) circle (0.4mm);
        \end{scope}
 \end{tikzpicture}
   \end{tabular}
    \end{subfigure}
    &
    \begin{subfigure}{0.25\linewidth}
        \begin{tikzpicture}
        \node[anchor=south west,inner sep=0] (image) at (0,0) {\includegraphics[ height=1\linewidth]{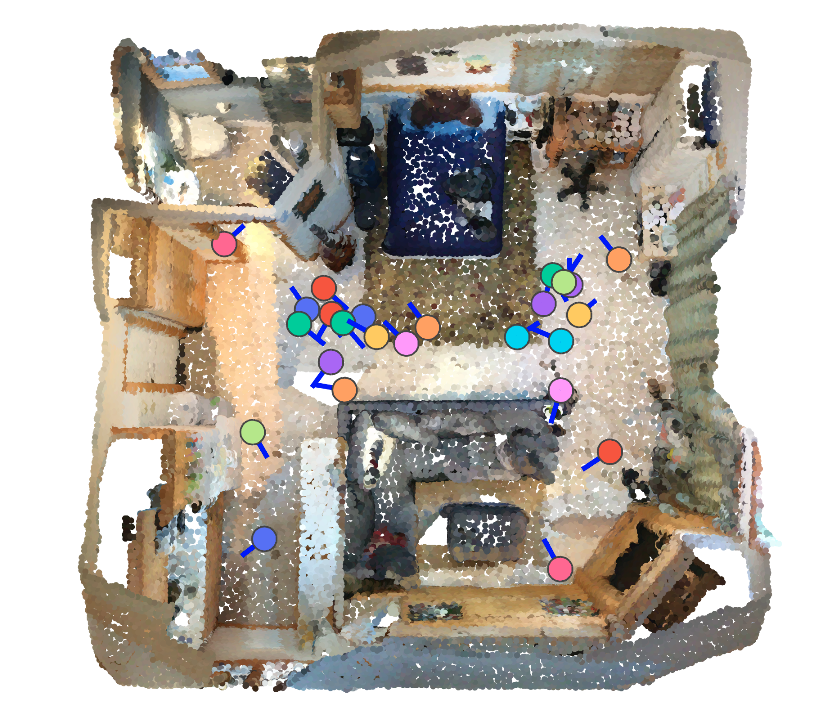}};
        \begin{scope}[x={(image.south east)},y={(image.north west)}]
        \fill[draw=none,fill=black!70!white, scale=1] (0.685,0.457) circle (0.2mm);
        \fill[draw=none,fill=black!70!white, scale=1] (0.7567,0.6374) circle (0.2mm);
        \fill[draw=none,fill=black!70!white, scale=1] (0.6855,0.526) circle (0.2mm);
        \end{scope}
 \end{tikzpicture}
   
   \end{subfigure}
   &
   \begin{subfigure}{0.11\linewidth}
    \begin{tabular}{c}
       \begin{tikzpicture}
        \node[anchor=south west,inner sep=0] (image) at (0,0) {\includegraphics[width=1\linewidth, height=0.65\linewidth]{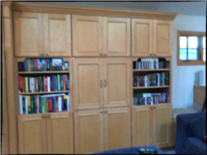}};
        \begin{scope}[x={(image.south east)},y={(image.north west)}]
        \draw[draw=blue, ultra thick] (0.1,0.8) -- (0.25,0.8);
        \definecolor{tempcolor}{rgb}{1,0.6,0.98}
        \fill[draw=black!70!white, thick, fill=tempcolor, scale=1] (0.1,0.8) circle (1.5mm);
        \fill[draw=none,fill=black!70!white, scale=1] (0.1,0.8) circle (0.4mm);
        \end{scope}
        \end{tikzpicture}
     \\
     \begin{tikzpicture}
        \node[anchor=south west,inner sep=0] (image) at (0,0) {\includegraphics[width=1\linewidth, height=0.65\linewidth]{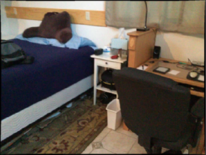}};
        \begin{scope}[x={(image.south east)},y={(image.north west)}]
        \draw[draw=blue, ultra thick] (0.1,0.8) -- (0.25,0.8);
        \definecolor{tempcolor}{rgb}{1,0.62,0.38}
        \fill[draw=black!70!white, thick, fill=tempcolor, scale=1] (0.1,0.8) circle (1.5mm);
        \fill[draw=none,fill=black!70!white, scale=1] (0.1,0.8) circle (0.4mm);
        \end{scope}
        \end{tikzpicture}
     \\
     \begin{tikzpicture}
        \node[anchor=south west,inner sep=0] (image) at (0,0) {\includegraphics[width=1\linewidth, height=0.65\linewidth]{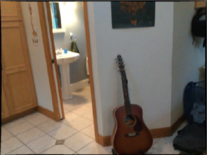}};
        \begin{scope}[x={(image.south east)},y={(image.north west)}]
        \draw[draw=blue, ultra thick] (0.1,0.8) -- (0.25,0.8);
        \definecolor{tempcolor}{rgb}{0,0.83,0.95}
        \fill[draw=black!70!white, thick, fill=tempcolor, scale=1] (0.1,0.8) circle (1.5mm);
        \fill[draw=none,fill=black!70!white, scale=1] (0.1,0.8) circle (0.4mm);
        \end{scope}
        \end{tikzpicture}
   \end{tabular}
   \end{subfigure}
    &
    \begin{subfigure}{0.27\linewidth}
   \begin{tabular}{c}
    \begin{tikzpicture}
        \node[anchor=south west,inner sep=0] (image) at (0,0) {\includegraphics[width=1\linewidth, height=0.45\linewidth]{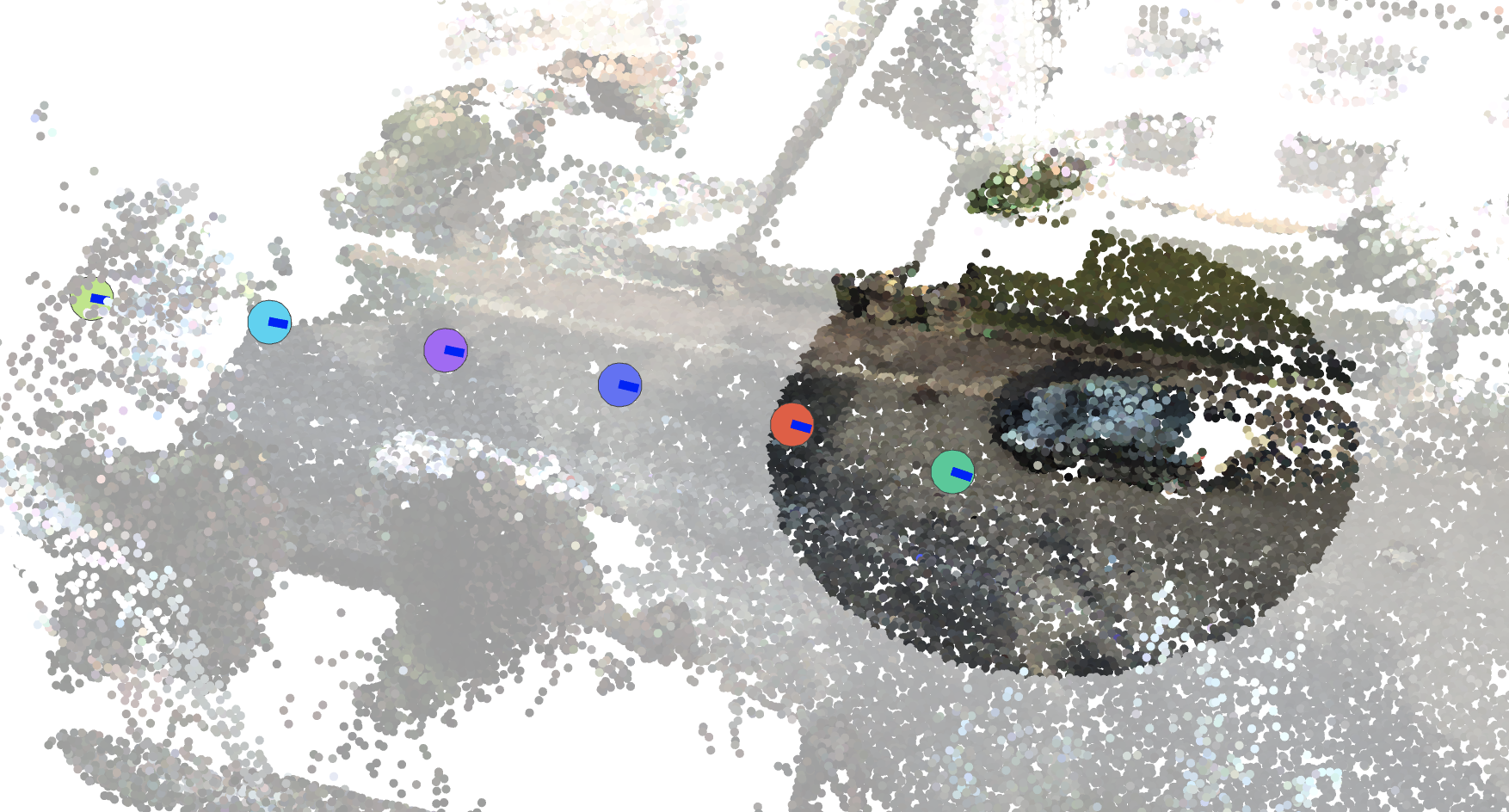}};
        \begin{scope}[x={(image.south east)},y={(image.north west)}]
        \fill[draw=none,fill=black!70!white, scale=1] (0.41,0.5225) circle (0.15mm);
        \end{scope}
 \end{tikzpicture}
   \\
   \begin{tikzpicture}
        \node[anchor=south west,inner sep=0] (image) at (0,0) {\includegraphics[width=1\linewidth, height=0.45\linewidth]{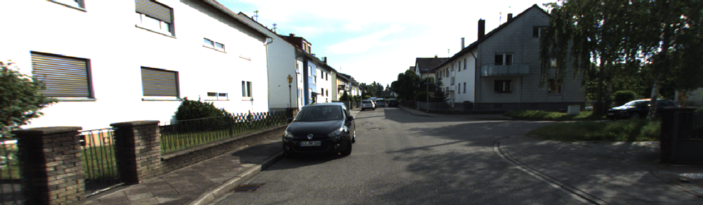}};
        \begin{scope}[x={(image.south east)},y={(image.north west)}]
        \draw[draw=blue, ultra thick] (0.05,0.9) -- (0.13,0.9);
        \definecolor{tempcolor}{rgb}{0.4,0.45,0.95}
        \fill[draw=black!70!white, thick, fill=tempcolor, scale=1] (0.05,0.9) circle (1.5mm);
        \fill[draw=none,fill=black!70!white, scale=1] (0.05,0.9) circle (0.4mm);
        \end{scope}
 \end{tikzpicture}
   \end{tabular}
    \end{subfigure}\\
    \begin{subfigure}{0.27\linewidth}
    \caption{S3DIS}
    \label{fig:dataset:s3dis}
    \end{subfigure}
    &
    \multicolumn{2}{c}{
    \begin{subfigure}{0.42\linewidth}
    \caption{ScanNet}
    \label{fig:dataset:scannet}
    \end{subfigure}
    }
    &
    \begin{subfigure}{0.27\linewidth}
    \caption{KITTI-360.}
    \label{fig:dataset:kitti}
    \end{subfigure}
    \end{tabular}
    \caption{{\bf Datasets.} Illustration of the sampling procedure for all considered datasets with point clouds alongside some of the available images. The 3D components of batches are constituted of spheres for \Subref{fig:dataset:s3dis} S3DIS, rooms for \Subref{fig:dataset:scannet} ScanNet, and cylinders for \Subref{fig:dataset:kitti} KITTI-360.}
    \label{fig:dataset}
\end{figure*}
\addtolength{\tabcolsep}{3pt}

\subsection{{Implementation Details}}
We use sparse encoding 
for mappings in order to only store compatible point-image pairs. This proves necessary for the large scale, in-the-wild setting with varying number of images seeing each point.
Our code is available at \code, the exact network and training configurations are given in the Appendix, and all the metrics of all runs can be accessed at \wandb.

\section{Experiments}

We propose several experiments on public large-scale semantic segmentation benchmarks to demonstrate the benefits of our deep multi-view aggregation module (DeepViewAgg). Our approach yields significantly better results than our 3D backbone directly operating on colorized point clouds. We set a new state-of-the-art for the highly contested S3DIS benchmark using only standard 2D and 3D architectures combined with our proposed module.
\subsection{Datasets}
\paragraph{S3DIS \cite{armeni20163d}.} This indoor dataset of office buildings contains over $278$ million semantically annotated 3D points across $6$ building areas---or \emph{folds}.
A companion dataset can be downloaded at \url{https://github.com/alexsax/2D-3D-Semantics}, and contains $1413$ equirectangular images.
To represent our large-scale, in-the-wild setting, we merge each fold into a large point cloud and discard all room-related information. We apply minor registration adjustments detailed in the Appendix.

\paragraph{ScanNet \cite{dai2017scannet}.} This indoor dataset contains over $1501$ scenes obtained from $2.5$ million RGB-D images with pose information. To account for the high redundancy between images, we select one in every $50$ image. This dataset deviates slightly from our intended setting as 2D and 3D are derived from the same sensors.

\paragraph{KITTI-360 \cite{Liao2021ARXIV}.} This large outdoor dataset contains over $100$k laser scans and $320$k images captured with a multi-sensor mobile platform. We use one image every five from the left perspective camera. We report the classwise performance on the official {withheld test} set.

\paragraph{General Setting.}  All datasets provide colorized point clouds obtained with dataset-specific preprocessings.
To handle the large size of scans, we define batches using a sampling strategy for S3DIS ($2$m-radius spheres) and KITTI-360 ($6$m-radius vertical cylinders), while we process ScanNet room-by-room, see \figref{fig:dataset}. We down-sample the point clouds for processing (S3DIS: 2cm, ScanNet: 3cm, KITTI-360: 5cm) and interpolate our prediction to full resolution for evaluation. To mitigate the memory impact of the 2D encoder, we also down-sample S3DIS images to $1024\times512$ but keep the full resolution for ScanNet ($320\times240$) and KITTI-360 ($1408\times376$).

\subsection{Quantitative Evaluation}
\begin{table}
    \caption{\textbf{Quantitative Evaluation.} Mean Intersection-over-Union of different state-of-the-art methods on S3DIS's Fold~5 and 6-fold, ScanNet Val, and KITTI-360 Test. All methods except the last line are trained on colorized point clouds. {\bf State-of-the-art}, \underline{ second highest}. ~$^1$ with 3D supervision only.}
    \label{tab:benchmark}
    \centering\small{
    \begin{tabular}{@{\extracolsep{0pt}}lllll}
        \multirow{2}{*}{Model} 
        &
        \multicolumn{2}{c}{S3DIS}
        &
        \multicolumn{1}{c}{ScanNet}
        &
        \multicolumn{1}{c}{KITTI}
        \\\cline{2-3}
        & Fold~5 & 6-Fold & \multicolumn{1}{c}{Val} & \multicolumn{1}{c}{360 Test}
        \\ \midrule 
        \multicolumn{5}{c}{\!\!\!\textit{Methods operating on colorized point clouds}}
        \\ \hline
        PointNet++ \cite{qi2017pointnetpp} &-&56.7 \cite{chaton2020torch} & 67.6~\cite{ptnpp} &  35.7~\cite{Liao2021ARXIV} \\
        SPG+SSP \cite{landrieu2018large, landrieu2019point} &61.7& 68.4 & - & -  \\
        MinkowskiNet \cite{choy20194d} & 65.4& 65.9\cite{chaton2020torch} & \underline{72.4}~\cite{nekrasov2021mix3d}& -  \\
        KPConv \cite{thomas2019kpconv} &67.1& 70.6 & 69.3~\cite{nekrasov2021mix3d}   & - \\
        RandLANet \cite{hu2020randla} &-& 70.0 & - & - \\
        PointTrans.\cite{engel2021point} &\bf 70.4 & \underline{73.5} & - & - \\
        Our 3D Backbone & 64.7 & 69.5 & 69.0 & \underline{53.9} \\
        \hline
        \multicolumn{5}{c}{\!\!\!\textit{Methods operating on point clouds and images}}
        \\ \hline
        MVPNet \cite{jaritz2019multi} & 62.4 & - & 68.3 & -\\
        VMVF \cite{kundu2020virtual} & 65.4&- &\bf 76.4 & - \\
        BPNet \cite{hu2021bidirectional} &-&-&69.7$^1$ &-\\
        3D Backbone+ &
        
        \multirow{2}{*}{\underline{67.2}} &
        \multirow{2}{*}{\textbf{74.7}} &
        \multirow{2}{*}{71.0} & 
        \multirow{2}{*}{\textbf{58.3}}
         \\
         DeepViewAgg \scriptsize{(ours)}\\
         \bottomrule
    \end{tabular}}
\end{table}

In \tabref{tab:benchmark}, we compare the performance of our approach and other learning methods on S3DIS, ScanNet Validation, and KITTI-360 Test using the classwise mean Intersection-over-Union (mIoU) as metric. Our method (DeepViewAgg) uses images in the 2D encoder and \emph{raw uncolored point clouds} in the 3D encoder. All other approaches, including our backbone (3D Backbone), use the colorized point clouds provided by the datasets.

DeepViewAgg sets a new state-of-the-art for S3DIS for all 6 folds and the second-highest performance for the 5th fold. In particular, we outperform the VMVF network \cite{kundu2020virtual}, showing that our multi-view aggregation model can overtake methods relying on costly virtual view generation using only available images. Furthermore, VMVF uses true depth maps, colorized point clouds, normals, and room-wise normalized information. In contrast, our method only uses raw XYZ data in the 3D encoder and estimates the mappings. 
Our approach also overtakes the recent PointTransformer \cite{engel2021point} (PointTrans.) by $1.2$ mIoU points, even though this method outperforms our 3D backbone by $4$ points on colorized points.
{Our model also improves the performance of our 3D backbone on the KITTI-360 test set by $4.4$ points, illustrating the importance of images for both indoor and outdoor datasets alike.}

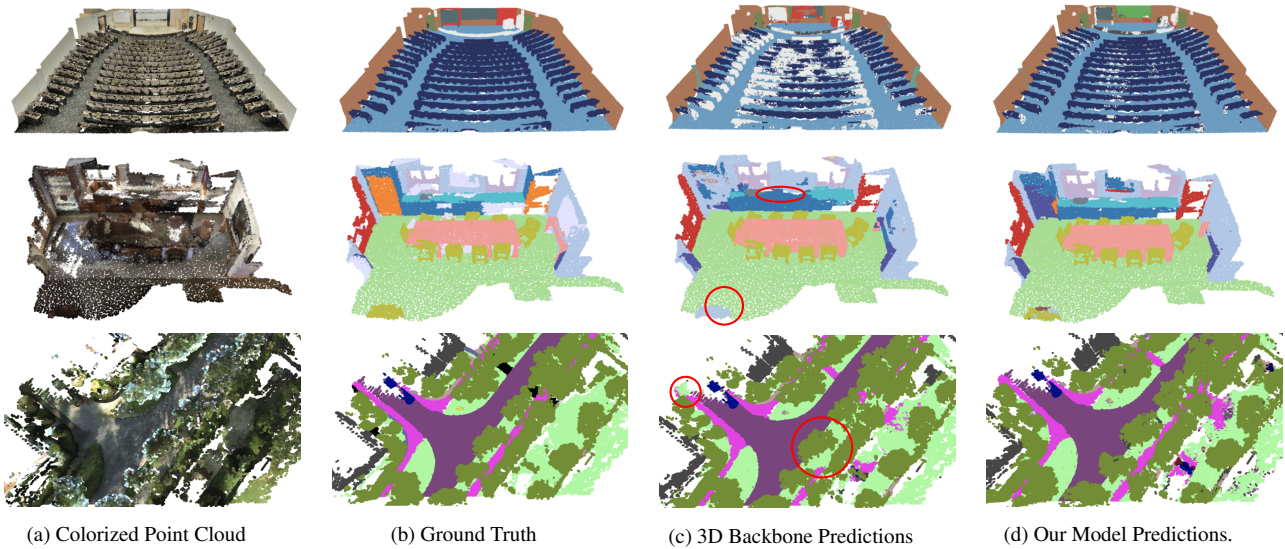
\begin{figure*}[t]
    \centering
    \begin{tabular}{cccc}
         \begin{subfigure}{0.225\textwidth}
         \begin{tabular}{c}
         \includegraphics[width=1\linewidth]{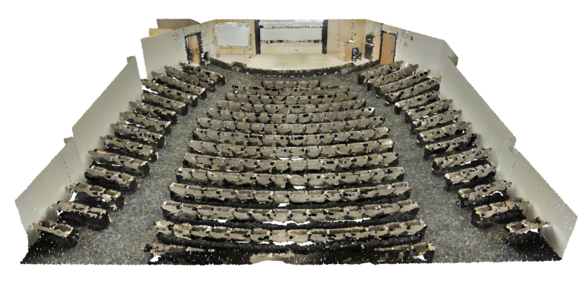}
         \\
         \scalebox{1}[.9]{\includegraphics[width=1\linewidth]{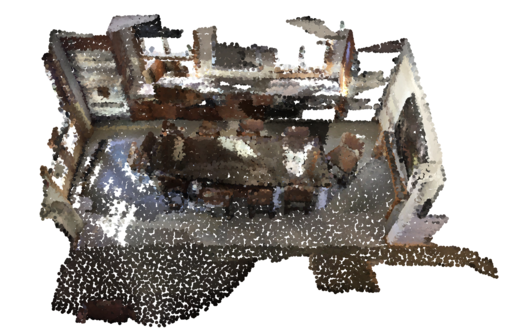}}
         \\
         \scalebox{1}[.9]{\includegraphics[width=1\linewidth]{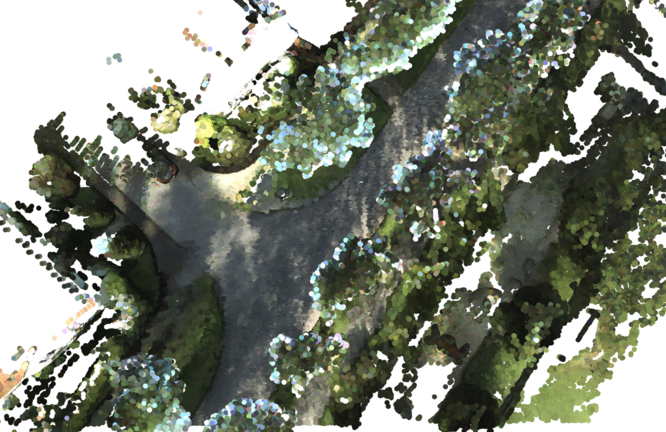}}
         \end{tabular}
         \caption{Colorized Point Cloud}
         \label{fig:quali:color}
         \end{subfigure}
         &
          \begin{subfigure}{0.225\textwidth}
         \begin{tabular}{c}
         \includegraphics[width=1\linewidth]{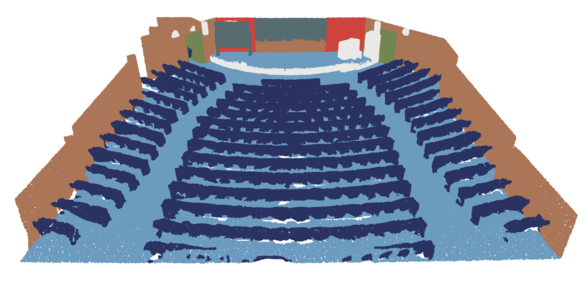}
         \\
         \scalebox{1}[.9]{\includegraphics[width=1\linewidth]{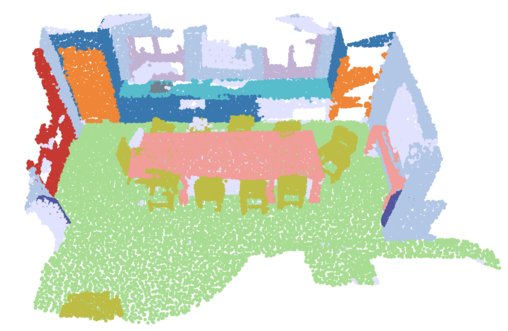}}
         \\
         \scalebox{1}[.9]{\includegraphics[width=1\linewidth]{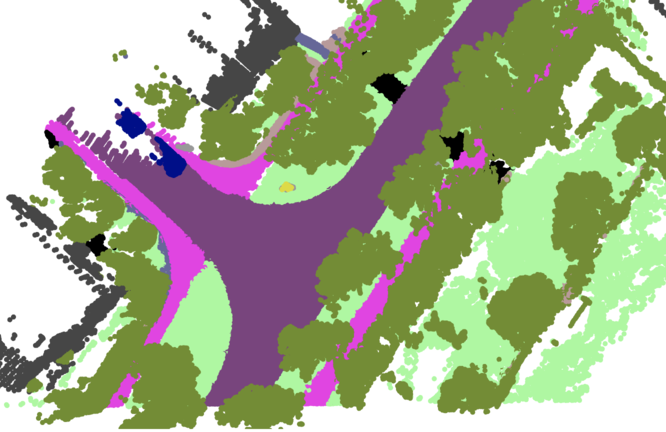}}
         \end{tabular}
         \caption{Ground Truth}
         \label{fig:quali:gt}
         \end{subfigure}
         &
          \begin{subfigure}{0.225\textwidth}
         \begin{tabular}{c}
         \includegraphics[width=1\linewidth]{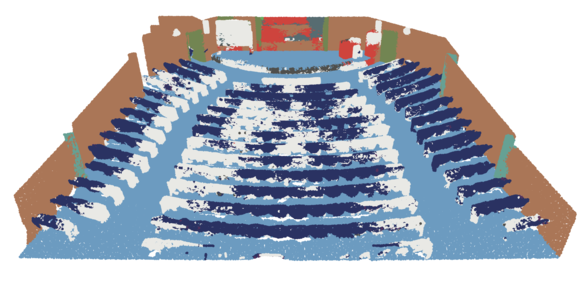}
         \\
          \begin{tikzpicture}
            \node[anchor=south west,inner sep=0] (image) at (0,0) { \scalebox{1}[.9]{\includegraphics[width=1\linewidth]{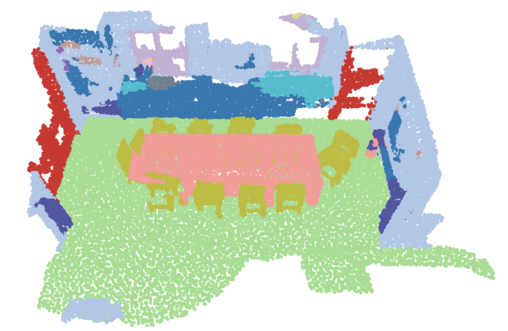}}};
            \begin{scope}[x={(image.south east)},y={(image.north west)}]
            \draw[draw=red, thick] (0.22,0.1) circle (2.5mm);
            \draw[draw=red, thick] (0.41,0.74) ellipse (3.2mm and 1mm);
            \end{scope}
          \end{tikzpicture}
         \\
          \begin{tikzpicture}
        \node[anchor=south west,inner sep=0] (image) at (0,0) { \scalebox{1}[.9]{\includegraphics[width=1\linewidth]{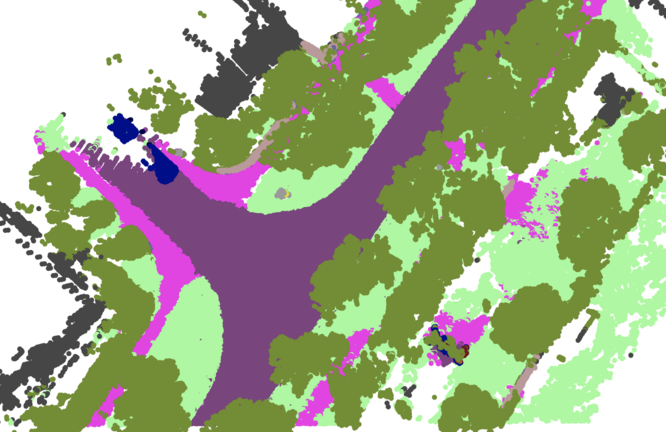}}};
        \begin{scope}[x={(image.south east)},y={(image.north west)}]
        \draw[draw=red, thick] (0.09,0.67) circle (2mm);
        \draw[draw=red, thick] (0.55,0.35) circle (4mm);
        \end{scope}
        \end{tikzpicture}
         \end{tabular}
         \caption{3D Backbone Predictions}
         \label{fig:quali:backbone}
         \end{subfigure}
         &
          \begin{subfigure}{0.225\textwidth}
         \begin{tabular}{c}
         \includegraphics[width=1\linewidth]{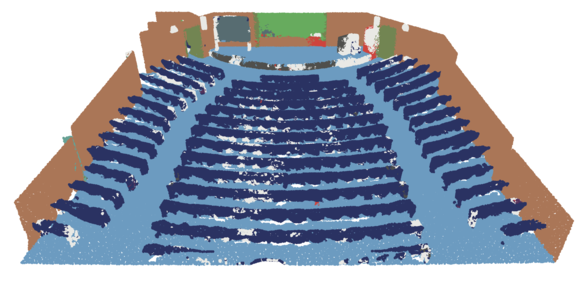}
         \\
         \scalebox{1}[.9]{\includegraphics[width=1\linewidth]{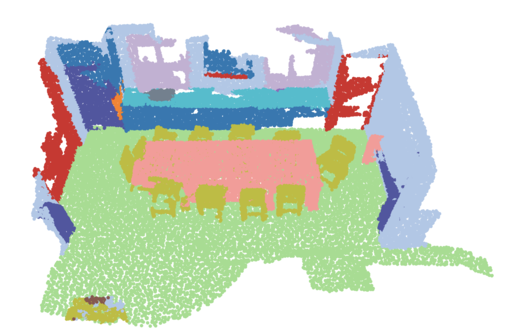}}
         \\
        \scalebox{1}[.9]{\includegraphics[width=1\linewidth]{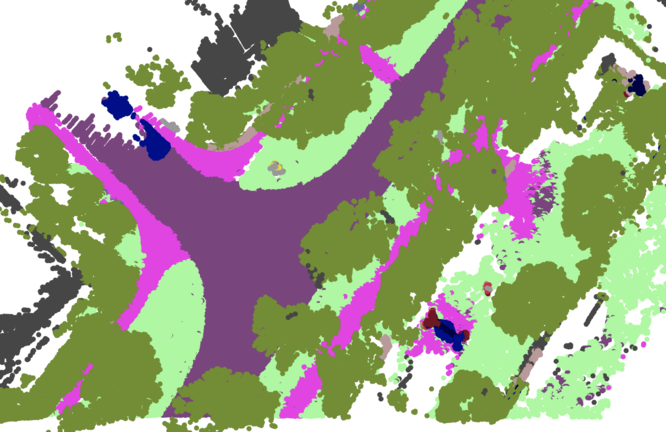}}
         \end{tabular}
         \caption{Our Model Predictions.}
         \label{fig:quali:ours}
         \end{subfigure}
    \end{tabular}
    \caption{{\bf Qualitative illustration.} Scenes from our considered datasets (top: S3DIS, middle: ScanNet, bottom: KITTI-360) with \Subref{fig:quali:color} colorized point clouds, \Subref{fig:quali:gt}
    ground truth point annotations, 
    \Subref{fig:quali:backbone}
    prediction of the backbone network operating on the colorized point cloud, and \Subref{fig:quali:ours} our method operating on raw uncolored point clouds and images.
    Our approach is able to use images to resolve cases in which the geometry is ambiguous or unusual, such as a large amphitheater with tiered rows of seats (top row).
    Color legend given in the Appendix.
    }
    \label{fig:quali}
\end{figure*}

While giving reasonable results, our method does not perform as well on the validation set of ScanNet comparatively.
We outperform the 2D/3D fusion method of BPNet \cite{hu2021bidirectional} when restricted to 3D annotations,  { illustrating the importance of view selection}. 
We argue that the limited variety in the camera points of view of ScanNet RGB-D scans, as well as their small field-of-view and blurriness reduce the quality of the information provided by images.
This is reinforced by the impressive performance of VMVF, which synthesizes its own images with controlled points of view and resolution.
{See \figref{fig:quali} for qualitative illustrations.}

\subsection{Analysis}
\begin{table}
    \caption{\textbf{Ablation Study.} Mean IoU comparison of different modalities and design choices on Fold~2 and Fold~5 of S3DIS down-sampled at 5cm for processing and KITTI-360 Val.}
    \label{tab:ablation}
    \centering
    \small{
    \begin{tabular}{@{\extracolsep{2pt}}lccc}
        
        \multirow{2}{*}{Model} & \multicolumn{2}{c}{S3DIS} & \multirow{1}{*}{KITTI} \\\cline{2-3}
        & Fold~2 & Fold~5 & 360 Val \\ \midrule 
        
        Best Configuration             & 63.2 & 67.5 & 57.8 \\\hline
        
        \multicolumn{4}{c}{\textit{Modality Combinations}} \\ \hline
        XYZRGB                         & -15.9 & -6.0 & -3.6 \\
        XYZ Average-RGB                & -10.8 & -7.0 & -4.9 \\
        XYZ                            & -19.5 & -9.5 & -4.1 \\
        Pure RGB                       & -5.3 & -5.4 & -14.5 \\
        Lower Image Resolution         & -5.9 & -0.8 & -0.7 \\
        Higher 3D Resolution           & -1.0 & -0.3 & - \\
        \hline
        
        \multicolumn{4}{c}{\textit{Design Choices}}        \\ \hline
        Late Fusion                    & -9.1 & -1.0 & -1.3 \\
        Only One Group                 & -4.8 & -0.8 & -0.4 \\
        No Gating                      & -3.0 & -0.4 & -1.1 \\
        No Dynamic Batch               & -6.9 & -1.9 & -4.5 \\
        No Pre-training                & -7.2 & -6.7 & -3.7 \\
        MaxPool                        & -0.8 & -1.5 & -2.9 \\
        Smaller 3D backbone            & -0.5 & -0.7 & +0.7 \\
        \bottomrule
    \end{tabular}}
\end{table}

We conduct further analyses on Fold~5 and Fold~2 of S3DIS (subsampled at 5cm for processing) {and the validation set of KITTI-360} in \tabref{tab:ablation}.
We added Fold~2 along the commonly used Fold~5, as it benefited most from our method, and hence is more conducive to evaluating the impact of our design choices.

\paragraph{Modality Combinations.} As observed in \tabref{tab:ablation}, combining a 3D deep network operating on raw 3D features and a 2D network with our method {(Best Configuration)} improves the performance by over $6$ to $15$ points compared to the same 3D backbone operating on colorized point clouds alone (XYZRGB). To illustrate that point colorization is not a trivial task, we train {our 3D backbone with point clouds colorized by averaging} for each point the color of all pixels in which it is visible (XYZ Average-RGB). Compared to the ``official'' colored point clouds, we observe a drop of $1$ point for Fold~5  and $1.3$ point for KITTI-360, but a gain of almost $5$ points for Fold~2. {This shows how different point cloud colorization schemes can yield vastly different results.}
Not using any radiometric information and purely relying on 3D {points without color} (XYZ) decreases the score of XYZRGB by a further $3$ to $4$ points on S3DIS. {For KITTI-360, XYZ outperforms XYZ Average-RGB, suggesting that poor colorization can even be detrimental.}

We also evaluate a scheme in which the 3D network is entirely removed, and 3D points are classified solely based on features coming from a 2D encoder-decoder and our view aggregation module, {without any 3D convolution} (Pure RGB). This method outperforms even {(XYZRGB)} for S3DIS, illustrating the relevance of images for point cloud segmentation. {On KITTI-360, as many 3D points are not seen by the cameras used, this approach perform worse.}

{Training our best 2D$+$3D model with images downsampled by a factor of $2$ (Lower Image Resolution) brings a large performance drop. In contrast, using $2$cm of 3D resolution instead of $5$cm (Higher 3D Resolution) has little impact for S3DIS.}
We conclude that when the images already contain fine-grained information, the impact of the resolution of the 3D voxel grid decreases.

\paragraph{Design Choices.} Using late fusion (Late Fusion) instead of early fusion gives comparable results on Fold~5 and KITTI-360, but significantly worse for Fold~2 for which the gain of using images is more pronounced.
Using only one feature group (Only One Group, $K=1$ in \equaref{eq:groups}) results in a drop of $4.8$ points for Fold~2, highlighting that our method can learn to treat different types of radiometric information specifically.
Removing the gating mechanism (No Gating, see \equaref{eq:gating}) decreases the IoU by $3$ points for Fold~2, and $1.1$ on KITTI-360.
Not using dynamic batches forces us to limit ourselves to $4$ full-size images per 3D sphere/cylinder, which results in performance drops of $2$ to $7$ points.
Pre-training the 2D network on related open-access datasets (No 2D Pre-Training) accounts for up to $7$ mIoU points. Not only do images contain rich radiometric information, but they also allow us to leverage the ubiquitous availability of annotated 2D datasets. 

Using featurewise max-pooling to merge the views results in a drop of $1$ to $1.5$ points for S3DIS {and $3$ points for KITTI-360}.
This illustrates that as long as we employ proper mapping, batching, and pre-training strategies, even simple pooling operations can perform very well. However, the addition of our model appears necessary to improve the precision even further and reach state-of-the-art results.

Switching our 3D backbone to a lighter version of MinkowskiNet with decreased widths, we observe no significant impact on the prediction quality. This suggests that we could use our approach successfully with smaller models.
{See the Appendix for further analysis of our design choices and of the influence of viewing conditions.}

\paragraph{Limitations.} While our method does not require sensors with aligned optical axes, true depth maps, or a meshing step, we still need camera poses. In some ``in the wild'' settings, they may not be available or require a pose estimation and registration step which may be costly and error-prone. 
Our mapping computation also relies on the assumption that the 2D and 3D modalities are acquired simultaneously.

Our multi-view aggregation method operates purely on viewing conditions and does not take the geometric and radiometric features into account in the computation of attention scores. We implemented a self-attention-based approach using such features, which resulted in a significant increase in memory usage without tangible benefits: the viewing conditions appear to be the most critical factor when selecting and aggregating images features.

\section{Conclusion}
We proposed a deep learning-based multi-view aggregation model for the semantic segmentation of large 3D scenes. Our approach uses the viewing condition of 3D points in images to select and merge the most relevant 2D features with 3D information. Combined with standard image and point cloud encoders, our method improves the state-of-the-art for two different datasets. 
Our full pipeline can run on a point cloud and a set of co-registered images at arbitrary positions without requiring colorization, meshing, or true depth maps.
These promising results illustrate the relevancy of using dedicated architectures for extracting information from images even for 3D scene analysis.

\ARXIV{}{\paragraph*{Acknowledgements} 
This work was funded by ENGIE Lab CRIGEN and carried on in the LASTIG research unit of Université Paris-Est.}

\FloatBarrier
\pagebreak
\pagebreak

\ARXIV{\paragraph*{Acknowledgements} 
This work was funded by ENGIE Lab CRIGEN and carried on in the LASTIG research unit of Université Paris-Est.
The authors wish to thank AI4GEO for sharing their computing resources. AI4GEO is a project funded by the French future investment program led by the Secretariat General for Investment and operated by public investment bank Bpifrance.
We thank Philippe Calvez, Dmitriy Slutskiy, Marcos Gomes-Borges, Gisela Lechuga, Romain Loiseau, Vivien Sainte Fare Garnot and Ewelina Rupnik for inspiring discussions and valuable feedback.}{}

{\small
\balance
\bibliographystyle{ieee_fullname}
\bibliography{mybib}
}

\ARXIV{
    \FloatBarrier
    \pagebreak
    \section*{\centering \LARGE Appendix}
    \setcounter{section}{0}
    \setcounter{figure}{0}
    \setcounter{table}{0}
    \renewcommand*{\theHsection}{appendix.\the\value{section}}
    \renewcommand\thefigure{\arabic{figure}}
    \renewcommand\thetable{\arabic{table}}
    \renewcommand\thefigure{SM-\arabic{figure}}
\renewcommand\thesection{SM-\arabic{section}}
\renewcommand\thetable{SM-\arabic{table}}
\renewcommand\theequation{SM-\arabic{equation}}
\renewcommand\thealgorithm{SM-\arabic{algorithm}} 

\section{Interactive Visualization and Code}
We release our code at \code  and our run metrics at \wandb. The provided code allows for reproduction of our experiments and inference using pretrained models.

Our repository also contains interactive visualizations as HTML files, showing different images and model predictions for spheres sampled in S3DIS, as shown in \figref{fig:interactivevisualization}. This tool makes it easier to see the additional insights brought by images than the visuals included in the paper.

\begin{figure}[ht]
    \centering
    \includegraphics[width=.8\linewidth]{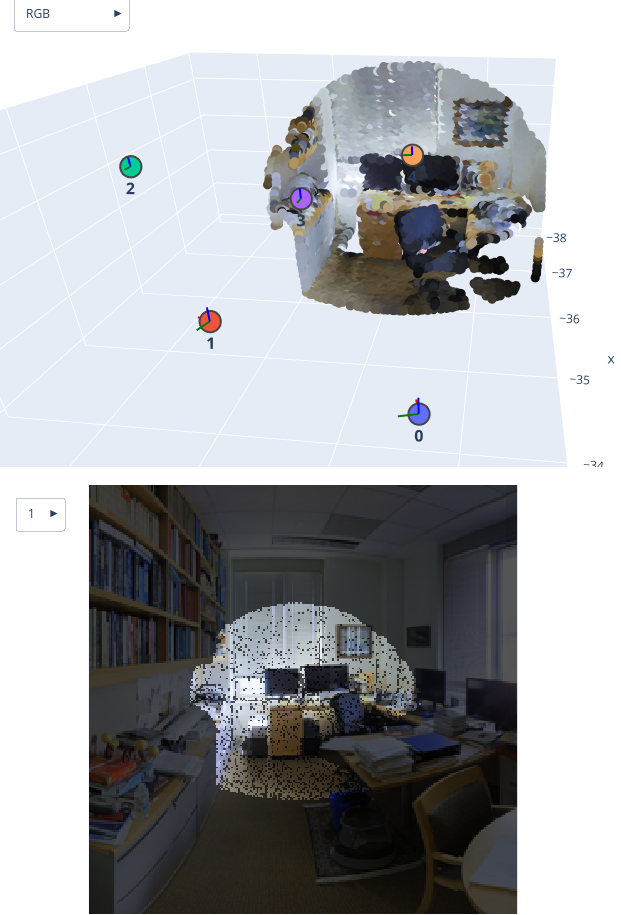}
    \caption{{\bf Visualizations.} We propose interactive visualizations of hybrid 2D/3D data along with predictions of our model. We also provide the code necessary to create more such visualizations.}
    \label{fig:interactivevisualization}
\end{figure}

\section{Efficient Point-Pixel Mapping }
In the Z-buffering step, we only consider points at a maximum distance $R=8$m for indoor scenes and $R=20$m for outdoor settings. We replace the points in image $i$ by cubes oriented towards $i$ and with a size given by the following formula involving $\dist(p,i)$ the distance between point $p$ and image $i$, $k=1$ a swell factor ruling how much closer cubes are expanded and $c$ the resolution of the voxel grid, or a typical inter-point distance (2-8cm in our experiments):
\begin{align}
    \text{size\_of\_cubes}(\dist(p,i)) = c(1 + k e^{-\dist(p,i) / R})~.
\end{align}
This heuristic increases the size of cubes that are close to the image to ensure that they do \emph{hide} the cubes behind them. Note that this heuristic operates on the size of the 3D cubes before camera projection and not on their projected pixel masks, which are computed based on camera intrinsic parameters. See Algorithm \ref{alg:zbuffer} for the pseudo-code of the mapping computation.

Storing point-pixel mappings for large-scale scenes with many images can be challenging. To minimize the memory impact of such a procedure, we use the Compressed Sparse Row (CSR) format. This allows us to represent the mappings compactly and treat large scenes at once. 

\begin{algorithm}
\caption{Z-buffering-Based Point-Pixel Mapping}\label{alg:zbuffer}
\begin{algorithmic}
\State{\textbf{Input:} $I$ image set, $P$ point cloud}
\For{$i \in I$}
    \State{zBuffer $\gets$ \texttt{maxFloat} array of size $i$ }
    \State{indexMap $\gets$ \texttt{NaN} array of size $i$ }
    \State{$P' \gets$ points of $P$ in frustum of $i$ and closer than $R$}
    \For{$p \in P'$}
        \State{$s \gets \text{size\_of\_cubes}(\dist(p,i))$}
        \State{$mask \gets $ pixel mask covered by the projection} 
        \State{\qquad  of a cube of size $s$ at $p$ onto the image $i$}
        \For{$(u, v) \in mask$}
            \If{$\dist(p,i) <$ zBuffer$[u, v]$}
                \State{zBuffer$[u, v] \gets \dist(p,i)$ }
                \State{indexMap$[u, v] \gets p$ }
            \EndIf
        \EndFor
    \EndFor
    \For{$p \in P'$}
    \If{$p$ appears in indexMap}
        \State{$\cM(p,i) \gets $ pixel at the projection of $p$ on $i$}
    \Else   
        \State{$\cM(p,i)$ not defined, $p$ not seen in $i$}
    \EndIf
    \EndFor
\EndFor
\end{algorithmic}
\end{algorithm}

\section{Projection Information}
We use $8$ handcrafted features to qualify the viewing conditions of a point $p$ seen in image $i$.
\begin{itemize}
    \item {\bf Normalized depth.} An image seeing a point at a distance may contain relevant contextual cues but poor textural information. We compute the distance $\dist(p,i)$ between point $p$ and image $i$ and divide by the maximum viewing distance $R=8$m for indoor scenes and $R=20$m for outdoor scenes.
    \item {\bf Local geometric descriptors.} The geometry of a point cloud can impact the quality of its views in images. Indeed, while planar surfaces may be better captured by a camera, a highly irregular surface may present many occlusions or grazing rays. We compute geometric descriptors (linearity, planarity, scattering) based on the eigenvalues of the covariance matrix between a point and its $50$ neighbors \cite{demantke2011dimensionality}.
    \item {\bf Viewing angle.} An image seeing a surface from a right angle may better capture its surroundings than if the view angle is slanted with respect to the surface. We compute the absolute value of the cosine between the viewing angle and the normal estimated from the covariance matrix calculated at the previous step.
    \item {\bf Pixel row.} To account for potential camera distortion near the top and bottom of the image (\eg for equirectangular images), we report the row of pixels and divide by the image height (number of rows). Note that we could derive a similar feature for cameras with radial distortion, such as fisheye cameras.
    \item {\bf Local density.} Density can impact occlusion and be an indicator of the local precision of the 3D sensor. We compute the area of the smallest disk containing the $50$th neighbor and normalize it by the square of the voxel grid resolution.
    \item {\bf Occlusion rate.} Occlusion may significantly impact the quality of the projected image features. We compute the ratio of the $50$ nearest neighbors of $p$ also seen in $i$.
\end{itemize}

\section{Fusion schemes}
We denote by $\{\f{2}_i\}_{i \in \II}$ a set of 2D feature maps of width $C$ associated with the images $\II$, typically obtained with a convolutional neural network. $\f{3}$ designates the raw feature of the point cloud given by the sensor: position, but also intensity/reflectance if available (not used in this paper). We denote by $\mathcal{P}(\f{2},\PP)$ the projection of the learned image features $\f{2}$ onto the point cloud $P$ by our multi-view aggregation technique. The early and late fusion schemes can be written as follows:
\begin{align}
    \label{eq:early}
    y_\text{early}&=\ClassTHREE \circ \DecTHREE \circ \EncTHREE
    \left(
        \left[
            \f{3},\mathcal{P}(\f{2},\PP)
        \right]
    \right)\\
    \label{eq:late}
    y_\text{late}&=\ClassTHREE 
    \left(
        \left[
                \DecTHREE \circ \EncTHREE
            \left(
                \f{3}
            \right)
            ,\mathcal{P}(\f{2},\PP)
        \right]
    \right)~.
\end{align}
For the intermediate fusion scheme, the 2D and 3D features are merged directly in the 3D encoder. Our 3D backbone follows a classic U-Net architecture, and its encoder is organized in $L$ levels  $\{\EncTHREE_l\}_{l=1}^L$ processing maps of increasingly coarse resolution. Each level $l>1$ is composed of of a downsampling module $\down_l$, typically strided convolutions ($\down_1=\text{Id}$), and a convolutional module $\conv_l$, typically a sequence of ResNet blocks. The 2D encoder is also composed of $L$ levels $\{\EncTWO_l\}_{l=1}^L$ corresponding to the different image resolutions. We propose to match the 2D and 3D levels at full resolution ($1024\times 512$ and $2$cm for S3DIS), and all subsequent levels after the same number of 2D/3D downsampling. At each level $l=1 \dots L$, we concatenate the downsampled higher resolution 3D map $\f{3}_{l-1}$ with the map $\f{2}_l$ obtained from the images at the matched resolution: 
\begin{align}
    \label{eq:inter}
   \f{3}_l &= \conv
   \left(
     \left[
      \down\left( \f{3}_{l-1}\right)
      ,
      \mathcal{P}(\f{2}_l,\PP)
     \right]
   \right)~,
\end{align}
with $\f{3}_0$ the raw 3D features. The decoder and classifiers follow the same organization than the 3D backbone.

\section{Dynamic-Size Image-Batching}
\begin{algorithm}[ht]
    \caption{Dynamic-Size Image-Batching}\label{alg:dynamic}
    \begin{algorithmic}
    \State{\textbf{Input:} $P_\text{sample}$ point cloud, $I$ image set}
    \State{\textbf{Parameters:} $B$ budget of pixels, $m$ border margin}
    \State{$I_\text{sample} \gets  \{i \mid i \in I \;\text{and}\; \exists p \in P_\text{sample} \;\text{s.t}\; i \in v(p)  \}$}
    \State{scores $\gets$ array of $0$ of size $I_\text{sample}$}
    \For{$i \in I_\text{sample}$}
        \State{$i \gets$ tightest crop($i$) to contain $P_\text{sample}$ with margin $m$}
        \State{scores$[i]\gets \score(i,P_\text{sample},I_\text{sample}, \varnothing)$}
    \EndFor
    \State{batch $\gets []$}
    \While{$B>0$ and length $I_\text{sample} > 0$}
        \State{pick $i \in I_\text{sample}$ randomly \wrt scores}
        \State{batch $\gets$ [batch, $i$]}
        \State{$I_\text{sample} \gets I_\text{sample} \setminus \{i\}$}
         \State{$B \gets B - \area(i)$}
         \For{$i \in I_\text{sample}$}
            \State{scores$[i]\gets \score(i,P_\text{sample},I_\text{sample}, \text{batch})$}
         \EndFor
    \EndWhile
    \end{algorithmic}
\end{algorithm}

We consider $P_\text{sample}$ a portion of a point cloud to add to the 3D part of a batch. In order to build the image batch we iteratively select images according to the following procedure. We first select the image set $I_\text{sample}$ seeing at least one point of $P_\text{sample}$.
For equirectangular images, we rotate the images to place the mappings at the center. We then crop each image $i$ along the tightest bounding box containing all seen point of $P_\text{sample}$ with a minimum margin of $m$ along a fixed set of image size along: $\text{crops}=\{64\times 64,128\times 64,128\times 128,256\times 128, 256\times 256, 512\times 256, 512\times 512, 1024\times 512\}$.
We then associate with each cropped image a score defined as follows:
\begin{align}
    &\score(i,P_\text{sample},I_\text{sample}, \text{batch}))=\\
    & \;\;\frac{\area(i)}{\max(\area(i) \mid i \in I_\text{sample})} \,+\\ 
    &\;\;\lambda \frac{\unseen(i,P_\text{sample},\text{batch}))}{\max(\unseen(i,P_\text{sample},\text{batch}) \mid i \in I_\text{sample})}~,
\end{align}
with $\area(i)$ the area of the cropped image $i$ in pixels, $\text{batch}$ the current image batch,   $\unseen(i,P_\text{sample},\text{batch})$ the number of points of $P_\text{sample}$ seen in image $i$ but not in any image of the current \text{batch}, and $\lambda=2$ a parameter controlling the trade-off between maximum area and maximum coverage. The current image batch is initialized as an empty set, but the scores must be updated as it is filled. The images are chosen randomly with a probability proportional to their score. We chose in all experiments a margin $m=8$ pixels and a budget corresponding to $4$ full resolution mapping.

\section{{Implementation Details}}
Our method is developed in Pytorch and is implemented within the open-source framework Torch-Points3d \cite{chaton2020torch}.

For our backbone 3D network, we use TorchPoint3D's \cite{chaton2020torch} Res16UNet34 implementation of MinkowskiNet \cite{choy20194d}. This UNet-like architecture comprises 5 encoding layers and 5 decoding layers. 
Encoding layers are composed of a strided convolution of \verb+kernel_size=[3, 2, 2, 2, 2]+ and \verb+stride=[1, 2, 2, 2, 2]+ followed by \verb+N=[0, 2, 3, 4, 6]+ ResNet blocks \cite{he2016identity} of channel size \verb+[128, 32, 64, 128, 256]+, respectively.
The decoding layers are built in the same manner, with a strided transposed convolution of \verb+kernel_size=[2, 2, 2, 2, 3]+ and \verb+stride=[2, 2, 2, 2, 1]+ as their first operation, followed by a concatenation with the corresponding skipped-features from the encoder and \verb+N=1+ ResNet block of channel size \verb+[128, 128, 96, 96, 96]+, respectively. A fully connected linear layer followed by a softmax converts the last features into class scores.
Unless specified otherwise, ReLU activation and batch normalization \cite{ioffe2015batch} are used across the architecture.

For our 2D encoders, we use the encoder part of pre-trained ResNet18 networks \cite{he2016deep}. For indoor scenes, we use the modified ResNet18 from \cite{csail} pre-trained on ADE20K \cite{zhou2017scene}, which has 5 layers of output channel sizes \verb+[128, 64, 128, 256, 512]+ and resolution \verb+scale=[4, 4, 8, 8, 8]+. The relatively high resolution  of the output feature map allows us to map the 2D features of size $C=512$ from the last layer directly to the point cloud without upsampling. For outdoor scenes, we use the ResNet18 from \cite{resnet18pretrained} pre-trained on Cityscapes \cite{Cordts2016Cityscapes}, which has 5 layers of output channel sizes \verb+[128, 64, 128, 256, 512]+ and resolution \verb+scale=[4, 4, 8, 16, 32]+. We use pyramid feature pooling \cite{zhao2017pyramid} on layers 1, 2, 3, and 4, which results in a feature vector of size $C=960$ passed to the mapped points.

For our DeepViewAgg module, the extracted image features are converted into view features of size $C$ with the MLP $\phi_0$ of size $C\mapsto  C\mapsto  C$. For the computation of quality scores, we use the following MLPs:  $\phi_1 : 8 \mapsto M  \mapsto M$, $\phi_2 : M \mapsto M  \mapsto M$, and $\phi_3 : 2M \mapsto M  \mapsto K$ with $M=32$ and $K=4$.

For indoor scenes, we use the lowest resolution map of a network \cite{csail} pretrained on ADE20K \cite{zhou2017scene}. 
For the outdoor scenes, we use pyramid feature pooling \cite{zhao2017pyramid} with a network \cite{resnet18pretrained} pretrained on Cityscapes \cite{Cordts2016Cityscapes}.
We use early fusion in all experiments unless specified otherwise.

All models are trained with SGD with an initial learning rate of $0.1$ for $200$ epochs with decay steps of $0.3$ at epoch $80,120,160$ for indoor datasets, and $20$ epochs with decay at epoch $10,16$ for the outdoor dataset.
{The pre-trained 2D networks use a learning rate $100$ times smaller than the rest of the model.}
We use random rotation and jittering for point clouds, random horizontal flip and color jittering for images, and featurewise jittering for descriptors of the viewing conditions.

For more details on the implementation, we refer the reader to our provided code.

\section{Supplementary Ablation}
{We propose further ablations whose results in \tabref{tab:supablation}.} {To assess the quality of our visibility model, we propose to compute the point-pixel mappings using the depth maps provided with S3DIS instead of Z-buffering.}
When running our method with such mappings (Mapping from Depth), we observe lower performances. This can be explained by the fact that depth-based mapping computation is sensitive to minor discrepancies between the depth map and the real point positions. Such a phenomenon can be observed on S3DIS depth images where the surfaces are viewed from a slanted angle, resulting in fewer point-image mappings being recovered. See \figref{fig:depth} for an illustration of this phenomenon.
In conclusion, not only can our approach bypass the need for specialized sensors or costly mesh reconstruction altogether, but our direct point-pixel mapping may yield better results than the provided mappings obtained with more involved methods.

\begin{table}[ht]
    \caption{\textbf{Supplementary Ablation Study.} Mean IoU comparison of different design choices on Fold~2 and Fold~5 of S3DIS down-sampled at 5cm for processing.}
    \label{tab:supablation}
    \centering
    \begin{tabular}{@{\extracolsep{2pt}}lll}
        \multirow{1}{*}{Model}         & Fold~2 & Fold~5 \\
        \midrule
        Best Configuration             & 63.1   &   67.5 \\
        \hline
        \multicolumn{3}{c}{\textit{Design Choices}}     \\
        \hline
        Mapping from Depth             &  -5.4  &  -1.9  \\
        Intermediate                   &  -7.6  &  -3.2  \\
        XYZRGB + DeepViewAgg           &  -0.5  &  -0.9  \\
        \bottomrule
    \end{tabular}
\end{table}

\begin{figure}[ht]
    \centering
    \begin{tabular}{c}
        \begin{subfigure}{\columnwidth}
            \includegraphics[width=\columnwidth]{figures/visibility_model/visibility_target.png}
            \caption{True depth.}
            \label{fig:depth:target}
        \end{subfigure}\\
        
        \begin{subfigure}{\columnwidth}
            \includegraphics[width=\columnwidth]{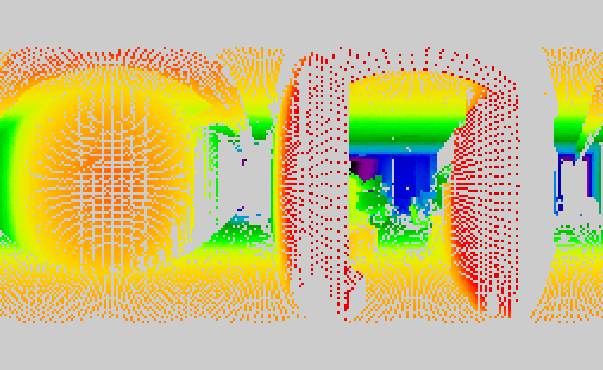}
            \caption{Depth-based visibility.}
            \label{fig:depth:computed}
        \end{subfigure}
    \end{tabular}
    \caption{{\bf Depth-Based Mapping Computation.} Based on an input depth map \Subref{fig:depth:target}, we compute the point-image mappings \Subref{fig:depth:computed} by searching points within a small margin of the target depth. We note that slight depth discrepancies near slanted surfaces prevents mapping from being recovered. Better seen on a monitor.}
    \label{fig:depth}
\end{figure}

To compare our fusion schemes, we evaluate a model with the intermediate fusion scheme described in \ref{eq:inter} (Intermediate). We observe that, for our module, intermediate fusion does not perform as well as early and late fusion. This could indicate that fusing modalities at their highest respective resolutions yields better results and that matching the encoder levels of 2D and 3D networks may not be straightforward. 
To ensure that our proposed module captures all radiometric information contained in colorized point clouds, we trained our chosen architecture to run on colorized point clouds \text{and} images (XYZRGB + DeepViewAgg). The resulting performance confirms that colorizing 3D points does not bring additional information not already captured by images.

\section{Influence of Maximum Depth}
{The maximum point-image depth is chosen as the distance beyond which adjacent 3D points appear in the same image pixel: $8$m for S3DIS sampled at $5$cm with images of width $1024$ and $20$m for KITTI-360. As illustrated in \tabref{tab:maxdepth}, reducing this parameter too much leads to a drop in performance both on S3DIS Fold~5 and KITTI-360, while slight modifications do not significantly affect the results. Since the number of point-image mappings grows quadratically with this parameter, one may consider smaller values to decrease the memory usage or computing time.}

\begin{table}[H]
    \caption{{\bf Effect of maximum Depth } We report the drop in mIoU when removing mappings beyond a threshold distance.}
    \label{tab:maxdepth}
    \centering
    \begin{tabular}{lc*{7}{c}}
        \multicolumn{8}{c}{S3DIS FOLD~5}\\
        \midrule
        Max depth & 8 & 7 & 6 & 5 & 4 & 3 & 2 \\
        mIoU drop & 0.0 & 0.0 & 0.0 & 0.4 & 0.6 & 1.9 & 8.6 \\
        \midrule~\\
    \end{tabular}
    \begin{tabular}{lc*{3}{c}}
        \multicolumn{4}{c}{KITTI-360 Val}\\
        \midrule
        Max depth & 20 & 15 & 10 \\
        mIoU drop & 0.0 & 0.5 & 2.6 \\
        \midrule~\\
    \end{tabular}
    \vspace{-5mm}
\end{table}

\section{Influence of Number of Images}
{
In contrast to existing methods (\eg MVPNet~\cite{jaritz2019multi}, VMVF~\cite{kundu2020virtual}, BPNet~\cite{hu2021bidirectional}), our set-based, sparse implementation of point-image mappings allows us to have a varying number of views per 3D point.
In \tabref{tab:numimages} we investigate the performance drop \wrt our best model when limiting the number of images per point cloud to a fixed number of images and not using dynamic batching. Under these conditions, our performance decreases by $6.8$ pts on S3DIS Fold~5 and $3.5$ pts on KITTI-360 when using only $3$ images, \ie the configuration of BPNet.
For comparison, 3D points of S3DIS are seen in $5.0$ images on average (STD $3.3$), and $2.5$ for KITTI-360 (STD $2.1$).
}

\begin{table}[ht]
    \caption{{Impact of the Number of Images. } We report the drop in mIoU when limiting the number of images per point cloud.}
    \label{tab:numimages} 
    \centering
    \begin{tabular}{lc*{6}{c}}
        \# images & 8 & 7 & 6 & 5 & 4 & 3 \\
        \midrule
        S3DIS Fold~5 & 0.5 & 1.2 & 1.7 & 2.1 & 3.5 & 6.8 \\
        KITTI-360 & 0.5 & 0.1 & 0.5 & 1.3 & 1.9 & 3.5 \\
        \midrule~\\
    \end{tabular}
    \vspace{-5mm}
\end{table}

\section{Influence of Viewing Conditions}
We propose to highlight the role viewing conditions descriptor. In \tabref{tab:viewablation}, we estimate the usage by our model of each feature as the drop in mIoU on S3DIS Fold~5 \& KITTI-360 when they are replaced by their dataset average (\eg all points appear at the same distance). We also measure the feature sensitivity by averaging the squared partial derivative \cite[3.3.1]{gevrey2003review} of the view compatibility score $x$ defined in (4) \wrt each view descriptor.
We observe that our model makes use of all observation features, and that the compatibility scores are most sensitive to small differences in scattering for S3DIS Fold~5, and depth for KITTI-360 Val.

\begin{table}[hb]
    \caption{{\bf Usage and Sensitivity of Viewing Conditions.} Feature usage is reported as a drop in mIoU, and the sensitivity is given as the proportion of squared partial derivative of the compatibility across all features.}
    \label{tab:viewablation}
    \centering \small
    \begin{tabular}{lcccc}
        \multirow{2}{*}{view feature} & \multicolumn{2}{c}{usage (mIoU drop)} & \multicolumn{2}{c}{sensitivity (in \%)} \\
         & S3DIS & KITTI-360 & S3DIS & KITTI-360 \\
        \midrule
        depth         & 1.1 & 1.7 & 12.6 & 46.0 \\
        linearity     & 1.0 & 0.8 & 11.9 &  0.7 \\
        planarity     & 1.0 & 1.4 & 15.8 &  1.9 \\
        scattering    & 0.7 & 1.0 & 52.7 &  0.7 \\
        viewing angle & 1.3 & 1.2 &  2.8 &  7.4 \\
        pixel row     & 1.1 & 0.8 &  1.6 & 33.2 \\
        local density & 1.2 & 1.3 &  0.6 &  1.8 \\
        occlusion     & 0.7 & 0.9 &  2.0 &  8.2 \\
        \bottomrule
    \end{tabular}
\end{table}

\begin{figure}
    \centering
    \includegraphics[width=\columnwidth]{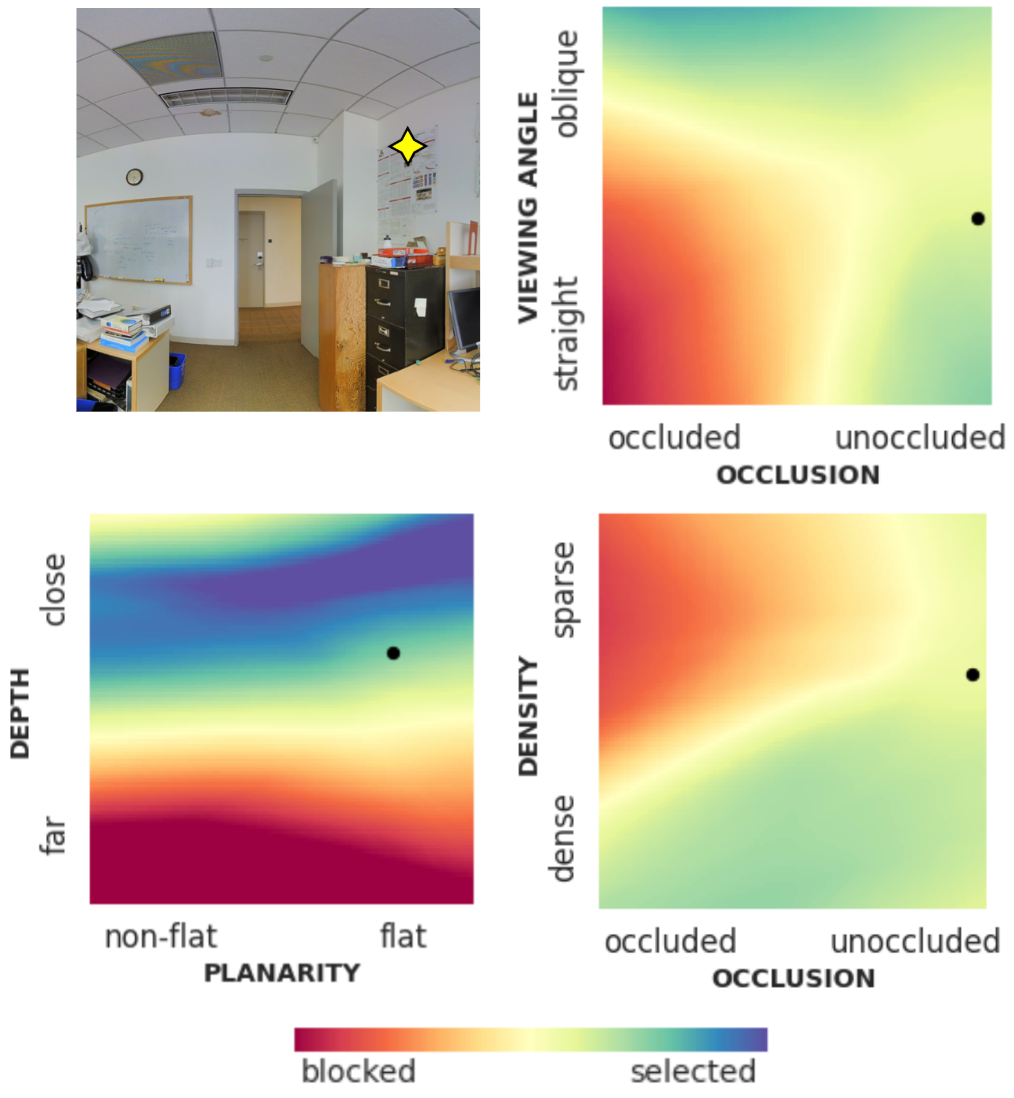}
    \caption{{\bf Influence of Viewing Descriptors.} Given a point-image pair \includegraphics[width=0.25cm]{images/star_3.png} (top left), we compute the quality scores when varying two of its viewing conditions from their initial values $\bullet$. For simplicity, we omit the influence of other images. We observe feature blocks specializing in retrieving information from views at a given depth range and containing planar objects (bottom left) or blocking straight yet occluded (top right) or sparse and occluded (bottom right) views.}
    \label{fig:viewing}
\end{figure}

To visualize the influence of viewing conditions, we represent in \figref{fig:viewing} quality score heatmaps when varying pairs of features for a given view point from S3DIS.

\section{S3DIS Adjustment}
We adjusted some room and image positions in S3DIS \cite{armeni20163d} and \url{https://github.com/alexsax/2D-3D-Semantics} to recover mappings between points and equirectangular images. More specifically, we rotate \verb+Area_2/hallway_11+ and \verb+Area_5/hallway_6+ by 180° around the Z-axis, and we shift and rotate all images in \verb+Area_5b+ by the same manually-found corrective offset and angle. These fixes are all available in our repository.

\section{Detailed Results}

\begin{table*}[t]
    \caption{{\bf Classwise Performance.} Classwise mIoU across all datasets for our 3D backbone network with and without learned multi-view aggregation. }
    \label{tab:classwise} 
    \begin{center}
    \footnotesize{
        \begin{tabular}{lc*{14}{c}}
            \multicolumn{15}{c}{S3DIS FOLD~5}\\
            Method & Avg & ceiling & floor & wall & beam & column & window & door & chair & table & bookcase & sofa & board & clutter\\
            \midrule
            3D Backbone   &         64.7 & \textbf{88.6} & \textbf{97.5} &        82.1 & \textbf{0.0} &         16.6 &          53.8 &          66.2 & \textbf{89.1} &        78.2 &         73.4 & \textbf{66.0} &       69.6 & \textbf{59.2}\\
            + DeepViewAgg & \textbf{67.2} &        87.2 &          97.3 & \textbf{84.3} & \textbf{0.0} & \textbf{23.4} & \textbf{67.6} & \textbf{72.6} &       87.8 & \textbf{81.0} & \textbf{76.4} &       54.9 & \textbf{82.4} &       58.7\\
            \midrule~\\
            \multicolumn{15}{c}{S3DIS 6-FOLD}\\
            3D Backbone   &         69.5 & \textbf{91.2} &        90.6 &          83.0 &          59.8 &          52.3 &          63.2 &          75.7 &          63.2 &          64.0 &          69.0 & \textbf{72.1} &       60.1 &          59.2\\
            + DeepViewAgg & \textbf{74.7} &        90.0 & \textbf{96.1} & \textbf{85.1} & \textbf{66.9} & \textbf{56.3} & \textbf{71.9} & \textbf{78.9} & \textbf{79.7} & \textbf{73.9} & \textbf{69.4} &        61.1 & \textbf{75.0} & \textbf{65.9}\\
            \midrule
        \end{tabular}\\~\\~\\
        
        \begin{tabular}{lc*{20}{c}}
            \multicolumn{22}{c}{ScanNet}\\
            Method & Avg & \rotatebox{90}{wall} & \rotatebox{90}{floor} & \rotatebox{90}{cabinet} & \rotatebox{90}{bed} & \rotatebox{90}{chair} & \rotatebox{90}{sofa} & \rotatebox{90}{table} & \rotatebox{90}{door} & \rotatebox{90}{window} & \rotatebox{90}{bookshelf} & \rotatebox{90}{picture} & \rotatebox{90}{counter} & \rotatebox{90}{desk} & \rotatebox{90}{curtain} & \rotatebox{90}{refrigerator} & \rotatebox{90}{shower cur.} & \rotatebox{90}{toilet} & \rotatebox{90}{sink} & \rotatebox{90}{bathtub} & \rotatebox{90}{other}\\
            \midrule
            3D Backbone   & 69.0 & \tiny 82.7 & \tiny \textbf{94.5} & \tiny \textbf{62.2} & \tiny 77.9 & \tiny 88.9 & \tiny \textbf{80.7} & \tiny 70.3 & \tiny 61.4 & \tiny 60.2 & \tiny \textbf{80.0} & \tiny 28.4 & \tiny 60.3 & \tiny \textbf{60.8} & \tiny 70.4 & \tiny 46.2 & \tiny 67.4 & \tiny \textbf{89.9} & \tiny 62.6 & \tiny \textbf{85.3} & \tiny 51.0\\
            + DeepViewAgg & \textbf{71.0} & \tiny \textbf{84.3} & \tiny 94.4 & \tiny 57.8 & \tiny \textbf{78.9} & \tiny \textbf{90.1} & \tiny 78.0 & \tiny \textbf{71.9} & \tiny \textbf{63.4} & \tiny \textbf{66.9} & \tiny 77.8 & \tiny \textbf{39.9} & \tiny \textbf{62.2} & \tiny 55.6 & \tiny \textbf{71.9} & \tiny \textbf{57.4} & \tiny \textbf{71.4} & \tiny 89.5 & \tiny \textbf{66.2} & \tiny 82.8 & \tiny \textbf{56.3}\\
            \midrule
        \end{tabular}\\~\\~\\
    
        \begin{tabular}{lc*{16}{c}}
            \multicolumn{17}{c}{KITTI-360 Val}\\
            Method & Avg & \rotatebox{90}{road} & \rotatebox{90}{sidewalk} & \rotatebox{90}{building} & \rotatebox{90}{wall} & \rotatebox{90}{fence} & \rotatebox{90}{pole} & \rotatebox{90}{traffic lig.} & \rotatebox{90}{traffic sig.} & \rotatebox{90}{vegetation} & \rotatebox{90}{terrain} & \rotatebox{90}{person} & \rotatebox{90}{car} & \rotatebox{90}{truck} & \rotatebox{90}{motorcycle} & \rotatebox{90}{bicycle}\\
            \midrule
            3D Backbone   & 54.2  & 90.6 & 74.4 & 84.5 & 45.3 & 42.9 & 52.7 & 0.5 & 38.6 & 87.6 & 70.3 & 26.9 & 87.3 & \textbf{66.0} & 28.2 & 17.2 \\
            + DeepViewAgg & \textbf{57.8} & \textbf{93.5} & \textbf{77.5} & \textbf{89.3} & \textbf{53.5} & \textbf{47.1} & \textbf{55.6} & \textbf{18.0} & \textbf{44.5} & \textbf{91.8} & \textbf{71.8} & \textbf{40.2} & \textbf{87.8} & 30.8 & \textbf{39.6} & \textbf{26.1} \\
            \midrule
        \end{tabular}\\
    }\end{center}
\end{table*}

We report in \tabref{tab:classwise} the classwise performance across all datasets. We see a clear improvement for indoor datasets for classes such as windows, boards, and pictures. These are expected results because these classes are hard to parse in 3D but easily identified in 2D. Besides, we can see that S3DIS's classes such as beams, columns, chairs, and tables also benefit from the contextual information provided by images. For the KITTI-360 dataset, the multimodal model outperforms the 3D-only baseline for all classes. We can see the benefit of image features on small objects or underrepresented classes in 3D, such as traffic signs, persons, trucks, motorcycles, and bicycles.

\balance

\begin{figure*}[!b]
    \centering
    \begin{tabular}{c}
        \large{S3DIS}
        \\\midrule
        \begin{tabular}{rlrlrlrlrl}
            \definecolor{tempcolor}{rgb}{0.91,0.90,0.41}
            \tikz \fill[fill=tempcolor, scale=0.3, draw=black] (0,0) rectangle (1,1);
            & \small{ceiling} 
            &
            \definecolor{tempcolor}{rgb}{.37,0.61,0.77}
            \tikz \fill[fill=tempcolor, scale=0.3, draw=black] (0,0) rectangle (1,1); 
            & \small{floor}
            &
            \definecolor{tempcolor}{rgb}{0.70,0.45,0.31}
            \tikz \fill[fill=tempcolor, scale=0.3, draw=black] (0,0) rectangle (1,1); 
            & \small{wall}
            &
            \definecolor{tempcolor}{rgb}{0.95,.58,0.51}
            \tikz \fill[fill=tempcolor, scale=0.3, draw=black] (0,0) rectangle (1,1);
            & \small{beam} &
            \definecolor{tempcolor}{rgb}{0.31,0.63,.58}
            \tikz \fill[fill=tempcolor, scale=0.3, draw=black] (0,0) rectangle (1,1);
            & \small{column} 
            \\
            \definecolor{tempcolor}{rgb}{0.30,0.68,.32}
            \tikz \fill[fill=tempcolor, scale=0.3, draw=black] (0,0) rectangle (1,1); 
            & \small{window}
            &
            \definecolor{tempcolor}{rgb}{.42,0.52,0.29}
            \tikz \fill[fill=tempcolor, scale=0.3, draw=black] (0,0) rectangle (1,1); 
            & \small{door}
            &
            \definecolor{tempcolor}{rgb}{.16,0.19,0.39}
            \tikz \fill[fill=tempcolor, scale=0.3, draw=black] (0,0) rectangle (1,1);
            & \small{chair}
            &
            \definecolor{tempcolor}{rgb}{.30,0.30,0.30}
            \tikz \fill[fill=tempcolor, scale=0.3, draw=black] (0,0) rectangle (1,1);
            & \small{table} 
            &
            \definecolor{tempcolor}{rgb}{.88,0.20,0.20}
            \tikz \fill[fill=tempcolor, scale=0.3, draw=black] (0,0) rectangle (1,1); 
            & \small{bookcase}
            \\
            \definecolor{tempcolor}{rgb}{0.35,.18,0.37}
            \tikz \fill[fill=tempcolor, scale=0.3, draw=black] (0,0) rectangle (1,1); 
            & \small{sofa}
            &
            \definecolor{tempcolor}{rgb}{.32,0.43,.45}
            \tikz \fill[fill=tempcolor, scale=0.3, draw=black] (0,0) rectangle (1,1);
            & \small{board}
            &
            \definecolor{tempcolor}{rgb}{0.91,0.91,.90}
            \tikz \fill[fill=tempcolor, scale=0.3, draw=black] (0,0) rectangle (1,1);
            & \small{clutter} 
            &
            \definecolor{tempcolor}{rgb}{0.,0.,0}
            \tikz \fill[fill=tempcolor, scale=0.3, draw=black] (0,0) rectangle (1,1);
            & \small{unlabeled} 
        \end{tabular}
        \\\midrule~\\
    
        \large{ScanNet}
        \\\midrule
        \begin{tabular}{rlrlrlrlrl}
            \definecolor{tempcolor}{rgb}{0.68, 0.78, 0.90}
            \tikz \fill[fill=tempcolor, scale=0.3, draw=black] (0,0) rectangle (1,1);
            & \small{wall} 
            &
            \definecolor{tempcolor}{rgb}{0.59, 0.87 , 0.54}
            \tikz \fill[fill=tempcolor, scale=0.3, draw=black] (0,0) rectangle (1,1); 
            & \small{floor}
            &
            \definecolor{tempcolor}{rgb}{0.12, 0.46, 0.70}
            \tikz \fill[fill=tempcolor, scale=0.3, draw=black] (0,0) rectangle (1,1); 
            & \small{cabinet}
            &
            \definecolor{tempcolor}{rgb}{1.   0.73, 0.47}
            \tikz \fill[fill=tempcolor, scale=0.3, draw=black] (0,0) rectangle (1,1);
            & \small{bed} &
            \definecolor{tempcolor}{rgb}{0.73 , 0.74, 0.13}
            \tikz \fill[fill=tempcolor, scale=0.3, draw=black] (0,0) rectangle (1,1);
            & \small{chair} 
            \\
            \definecolor{tempcolor}{rgb}{0.54, 0.33 , 0.29}
            \tikz \fill[fill=tempcolor, scale=0.3, draw=black] (0,0) rectangle (1,1); 
            & \small{sofa}
            &
            \definecolor{tempcolor}{rgb}{1.   0.59, 0.58}
            \tikz \fill[fill=tempcolor, scale=0.3, draw=black] (0,0) rectangle (1,1); 
            & \small{table}
            &
            \definecolor{tempcolor}{rgb}{0.83, 0.15, 0.15}
            \tikz \fill[fill=tempcolor, scale=0.3, draw=black] (0,0) rectangle (1,1);
            & \small{door}
            &
            \definecolor{tempcolor}{rgb}{0.77, 0.69, 0.83}
            \tikz \fill[fill=tempcolor, scale=0.3, draw=black] (0,0) rectangle (1,1); 
            & \small{window}
            &
            \definecolor{tempcolor}{rgb}{0.58, 0.40, 0.74}
            \tikz \fill[fill=tempcolor, scale=0.3, draw=black] (0,0) rectangle (1,1); 
            & \small{bookshelf}
            \\
            \definecolor{tempcolor}{rgb}{0.76, 0.61, 0.58}
            \tikz \fill[fill=tempcolor, scale=0.3, draw=black] (0,0) rectangle (1,1);
            & \small{picture}
            &
            \definecolor{tempcolor}{rgb}{0.09, 0.74, 0.81}
            \tikz \fill[fill=tempcolor, scale=0.3, draw=black] (0,0) rectangle (1,1);
            & \small{counter} 
            &
            \definecolor{tempcolor}{rgb}{0.96, 0.71, 0.82}
            \tikz \fill[fill=tempcolor, scale=0.3, draw=black] (0,0) rectangle (1,1);
            & \small{desk} 
            &
            \definecolor{tempcolor}{rgb}{0.85, 0.85, 0.55}
            \tikz \fill[fill=tempcolor, scale=0.3, draw=black] (0,0) rectangle (1,1);
            & \small{curtain}
            &
            \definecolor{tempcolor}{rgb}{1.   0.49, 0.05}
            \tikz \fill[fill=tempcolor, scale=0.3, draw=black] (0,0) rectangle (1,1); 
            & \small{refrigerator}
            \\
            \definecolor{tempcolor}{rgb}{0.61, 0.85, 0.89}
            \tikz \fill[fill=tempcolor, scale=0.3, draw=black] (0,0) rectangle (1,1);
            & \small{shower curtain}
            &
            \definecolor{tempcolor}{rgb}{0.17, 0.62, 0.17}
            \tikz \fill[fill=tempcolor, scale=0.3, draw=black] (0,0) rectangle (1,1);
            & \small{toilet} 
            &
            \definecolor{tempcolor}{rgb}{0.43, 0.50, 0.56}
            \tikz \fill[fill=tempcolor, scale=0.3, draw=black] (0,0) rectangle (1,1);
            & \small{sink} 
            &
            \definecolor{tempcolor}{rgb}{0.89, 0.46, 0.76}
            \tikz \fill[fill=tempcolor, scale=0.3, draw=black] (0,0) rectangle (1,1);
            & \small{bathtub}
            &
            \definecolor{tempcolor}{rgb}{0.32, 0.32, 0.63}
            \tikz \fill[fill=tempcolor, scale=0.3, draw=black] (0,0) rectangle (1,1);
            & \small{otherfurniture} 
            \\
            \definecolor{tempcolor}{rgb}{0.88, 0.88, 1.  }
            \tikz \fill[fill=tempcolor, scale=0.3, draw=black] (0,0) rectangle (1,1);
            & \small{ignored}
        \end{tabular}\\
        \midrule~\\
        
        \large{KITTI-360}
        \\\midrule
        \begin{tabular}{rlrlrlrlrl}
            \definecolor{tempcolor}{rgb}{0.50, 0.25, 0.50}
            \tikz \fill[fill=tempcolor, scale=0.3, draw=black] (0,0) rectangle (1,1);
            & \small{road} 
            &
            \definecolor{tempcolor}{rgb}{0.95, 0.13, 0.90}
            \tikz \fill[fill=tempcolor, scale=0.3, draw=black] (0,0) rectangle (1,1);
            & \small{sidewalk} 
            &
            \definecolor{tempcolor}{rgb}{0.27, 0.27, 0.27}
            \tikz \fill[fill=tempcolor, scale=0.3, draw=black] (0,0) rectangle (1,1);
            & \small{building} 
            &
            \definecolor{tempcolor}{rgb}{0.4 , 0.4 , 0.61}
            \tikz \fill[fill=tempcolor, scale=0.3, draw=black] (0,0) rectangle (1,1);
            & \small{wall} 
            &
            \definecolor{tempcolor}{rgb}{0.74, 0.6 , 0.6 }
            \tikz \fill[fill=tempcolor, scale=0.3, draw=black] (0,0) rectangle (1,1);
            & \small{fence} 
            \\
            \definecolor{tempcolor}{rgb}{0.6 , 0.6 , 0.6 }
            \tikz \fill[fill=tempcolor, scale=0.3, draw=black] (0,0) rectangle (1,1);
            & \small{pole} 
            &
            \definecolor{tempcolor}{rgb}{0.98, 0.66, 0.11}
            \tikz \fill[fill=tempcolor, scale=0.3, draw=black] (0,0) rectangle (1,1);
            & \small{traffic light} 
            &
            \definecolor{tempcolor}{rgb}{0.86, 0.86, 0.  }
            \tikz \fill[fill=tempcolor, scale=0.3, draw=black] (0,0) rectangle (1,1);
            & \small{traffic sign} 
            &
            \definecolor{tempcolor}{rgb}{0.41, 0.55, 0.13}
            \tikz \fill[fill=tempcolor, scale=0.3, draw=black] (0,0) rectangle (1,1);
            & \small{vegetation} 
            &
            \definecolor{tempcolor}{rgb}{0.59, 0.98, 0.59}
            \tikz \fill[fill=tempcolor, scale=0.3, draw=black] (0,0) rectangle (1,1);
            & \small{terrain} 
            \\
            \definecolor{tempcolor}{rgb}{0.86, 0.07, 0.23}
            \tikz \fill[fill=tempcolor, scale=0.3, draw=black] (0,0) rectangle (1,1);
            & \small{person} 
            &
            \definecolor{tempcolor}{rgb}{0.  , 0.  , 0.55}
            \tikz \fill[fill=tempcolor, scale=0.3, draw=black] (0,0) rectangle (1,1);
            & \small{car} 
            &
            \definecolor{tempcolor}{rgb}{0.  , 0.  , 0.27}
            \tikz \fill[fill=tempcolor, scale=0.3, draw=black] (0,0) rectangle (1,1);
            & \small{truck} 
            &
            \definecolor{tempcolor}{rgb}{0.  , 0.  , 0.90}
            \tikz \fill[fill=tempcolor, scale=0.3, draw=black] (0,0) rectangle (1,1);
            & \small{motorcycle} 
            &
            \definecolor{tempcolor}{rgb}{0.46, 0.04, 0.12}
            \tikz \fill[fill=tempcolor, scale=0.3, draw=black] (0,0) rectangle (1,1);
            & \small{bicycle} 
            \\
            \definecolor{tempcolor}{rgb}{0.  , 0.  , 0.  }
            \tikz \fill[fill=tempcolor, scale=0.3, draw=black] (0,0) rectangle (1,1);
            & \small{ignored} 
            &
        \end{tabular}\\
        \midrule
    \end{tabular}
    \caption{{\bf Color Legend.} 
    }
    \label{fig:colorlegend}
\end{figure*}

}{}

\end{document}